\documentclass[reqno]{amsart}
\usepackage{graphicx}
\usepackage{amsmath,amssymb,amsthm} 
\usepackage{bm}
\usepackage{color}
\usepackage[scriptsize,tight]{subfigure}
\usepackage{mathabx}
\usepackage[font=small]{caption}
\usepackage{epsfig}
\usepackage{rotating}
\usepackage{tabularx}

\DeclareMathOperator{\rank}{rank}

\DeclareMathOperator{\SE}{SE}

\DeclareMathOperator{\SO}{SO}

\newcommand{\T}{^{\top}}
\newcommand{\nn}{\hat{\bm{n}}}
\newcommand{\nd}{\nn_{\mathrm{m}}}

\usepackage{ctable}
\newcommand{\otoprule}{\midrule[\heavyrulewidth]}

\title[Cavlectometry]{Cavlectometry: Towards Holistic Reconstruction of Large Mirror Objects}
\author{J. Balzer, D. Acevedo-Feliz, S. Soatto, S. H\"ofer, M. Hadwiger, and J. Beyerer}
\begin{document}

\maketitle

\begin{abstract}
We introduce a method based on the deflectometry principle for the reconstruction of specular objects exhibiting significant size and geometric complexity. A key feature of our approach is the deployment of an \emph{Automatic Virtual Environment} (CAVE) as pattern generator. To unfold the full power of this extraordinary experimental setup, an optical encoding scheme is developed which accounts for the distinctive topology of the CAVE. Furthermore, we devise an algorithm for detecting the object of interest in raw deflectometric images. The segmented foreground is used for single-view reconstruction, the background for estimation of the camera pose, necessary for calibrating the sensor system. Experiments suggest a significant gain of coverage in single measurements compared to previous methods. To facilitate research on specular surface reconstruction, we will make our data set publicly available.
\end{abstract}

\section{Introduction}\label{sec:introduction}
\subsection{Motivation}\label{subsec:motivation}
Reconstructing the geometry of an object or scene from a collection of images is an important problem in applications ranging from assisted surgery to additive manufacturing, quality control, augmented reality, just to mention a few. The problem is particularly challenging when objects exhibit complex non-Lambertian reflectance. For instance, specular surfaces cannot be reconstructed without \emph{some} assumption on the surrounding environment. While generic prior assumptions may be sufficient for classification purposes, accurate metrology calls for precise knowledge of the surroundings, that can be thought of as the ``illumination''. In \emph{deflectometry}, a controllable illuminant (e.g., a liquid-crystal display, LCD) is reflected through the object's surface onto the image plane, see Fig.~\ref{fig:classical_setup}. In order to reconstruct a model of the unknown surface from the known illuminant (input) and its measured image output, the input must be sufficiently exciting. Typically, a series of optical code words is displayed, identifying each of the points on the illuminant uniquely. This way, dense correspondence between the domains of input and output signal is established, the so-called \emph{light map}, assigning to a pixel the scene point\footnote{The terms scene, illuminant, and pattern generator are used synonymously throughout this paper.} it sees \emph{via} the specular surface~\cite{Donner2004}. 

\begin{figure}[b]
\centering
\subfigure[]{\label{fig:adato}\includegraphics[height=3.3cm]{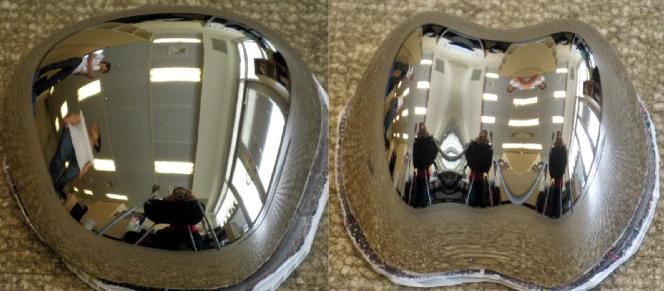}\hspace{0.05cm}\tiny\begin{turn}{90}(Courtesy of O. Ben-Shahar)\end{turn}}\hfill
\subfigure[]{\label{fig:gravy_boat}\includegraphics[height=3.3cm]{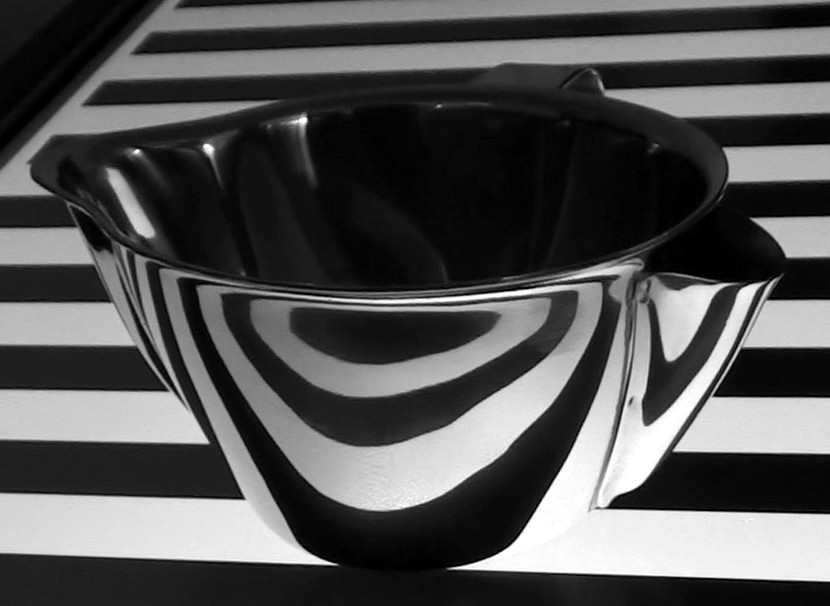}\hspace{0.05cm}\tiny\begin{turn}{90}(Courtesy of M. Liu)\end{turn}}\\
\subfigure[]{\label{fig:bunny_weinmann}\includegraphics[height=4.5cm]{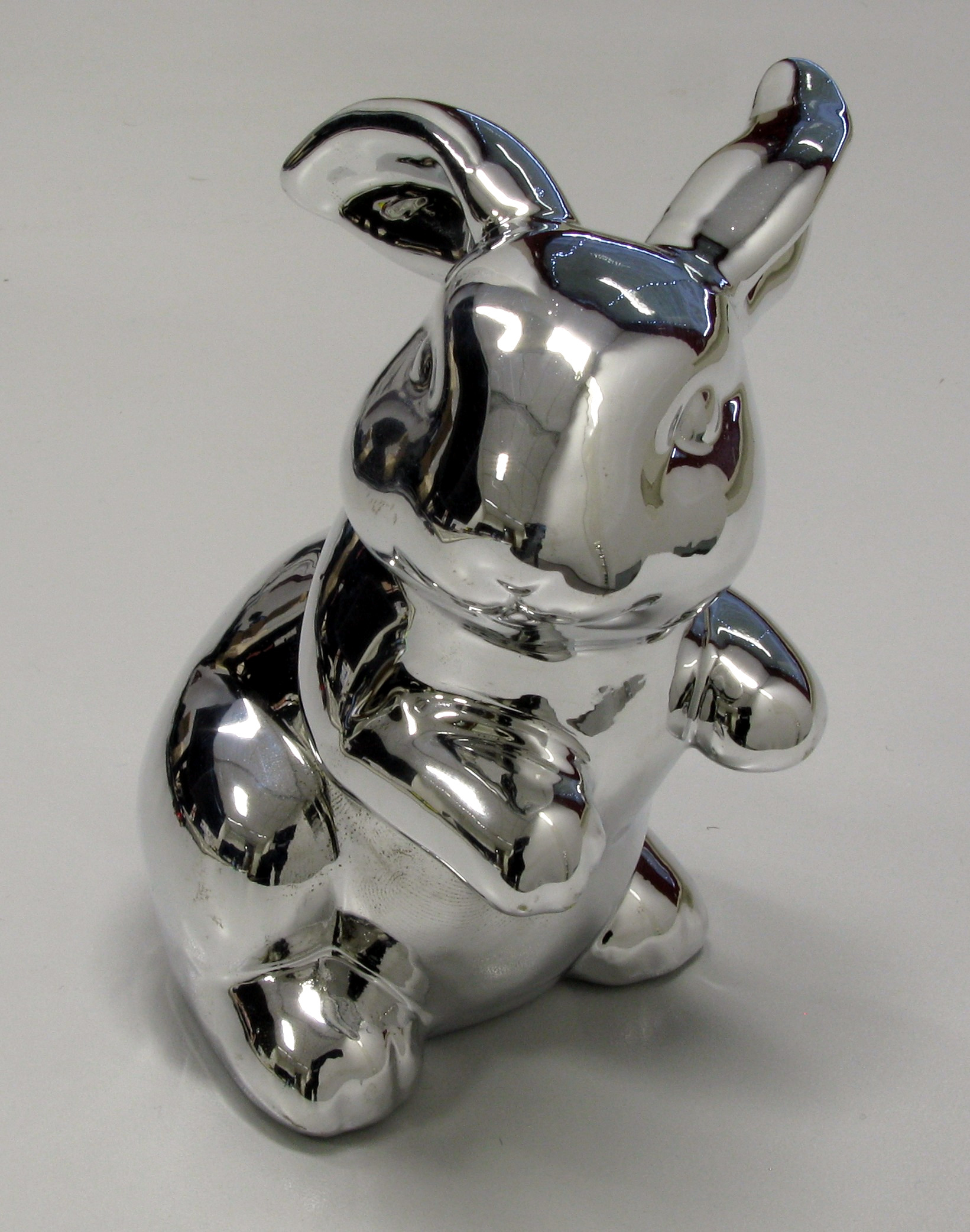}\hspace{0.05cm}\tiny\begin{turn}{90}(Courtesy of M. Weinmann)\end{turn}}\hfill
\subfigure[]{\label{fig:hood_mv}\includegraphics[height=4.5cm]{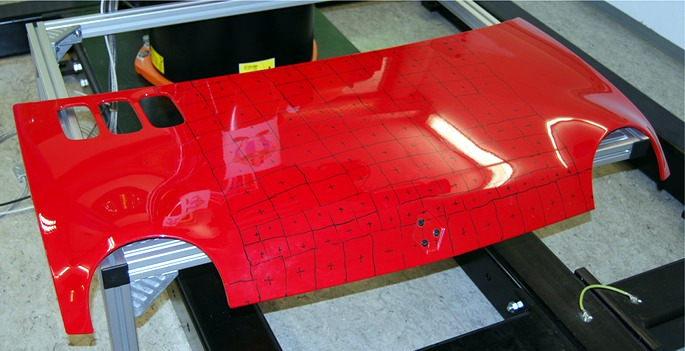}}
\caption{Objects of study in some of the current literature on mirror surface reconstruction~\cite{Adato2010,Liu2013,Weinmann2013,Balzer2011}: \subref{fig:adato} Convex surface patches compress the reflected image of the environment. \subref{fig:gravy_boat} - \subref{fig:bunny_weinmann} This  effect can be remedied by pla\-cing the surface close to the pattern generator and sampling a large number of vantage points. \subref{fig:hood_mv} The markings on this engine hood give an impression of how many measurements were required by the method described in~\cite{Balzer2011}.}
\label{fig:setup}
\end{figure}

To make the basic deflectometry procedure viable for industrial applications, an important open problem needs to be solved: Convex objects, no matter how small, act as de-magnifying lenses and thus reflect a significant portion of the environment. For instance, in Fig.~\ref{fig:adato}, almost half of the scene becomes visible in a spherical patch of just a few square centimeters. There are essentially two countermeasures: One can either decompose the surface into smaller tractable pieces, cf.~\cite{Balzer2011}, but depending on the geometry of the object, the number of required views can easily reach the hundreds. Alternatively, many state-of-the-art methods optimize the effective area on the mirror surface by pla\-cing the pattern generator in close proximity, cf. Fig.~\ref{fig:gravy_boat}. Going one step further, Hong et al.~\cite{Hong2009} arrange five LCDs tightly around the object leaving a small gap in between for a \emph{single} camera to peek through. Both strategies, however, make it difficult if not impossible to account for occlusions, the crux of any reconstruction methodology. In fact, one of the idiosyncrasies of deflectometry is that it may suffer from occlusions not just of the surface itself but also of the \emph{illuminant}, i.e., light rays can be blocked \emph{after} interacting with the surface \emph{and before}. For the purpose of reconstruction, each point on the surface must grant at least one unobstructed view on the illuminant, and ideally, images should be acquired from different camera positions until this condition is completely satisfied. To this end, Weinmann et al.~\cite{Weinmann2013} furnish their system with some of the desired flexibility by combining eleven cameras and three LCDs with a turntable to carry the specimen. While their approach yields impressive reconstructions of small-scale objects, first, it still relies on a very large number of measurements, $792$ for the example in Fig.~\ref{fig:bunny_weinmann}, and second, it excludes large surfaces, e.g. automobile parts, see Fig.~\ref{fig:hood_mv}, which are certainly of interest to the practitioner.

\subsection{Contribution and overview}
\begin{figure}[t]
\centering
\subfigure[]{\label{fig:classical_setup}\includegraphics[width=0.49\columnwidth]{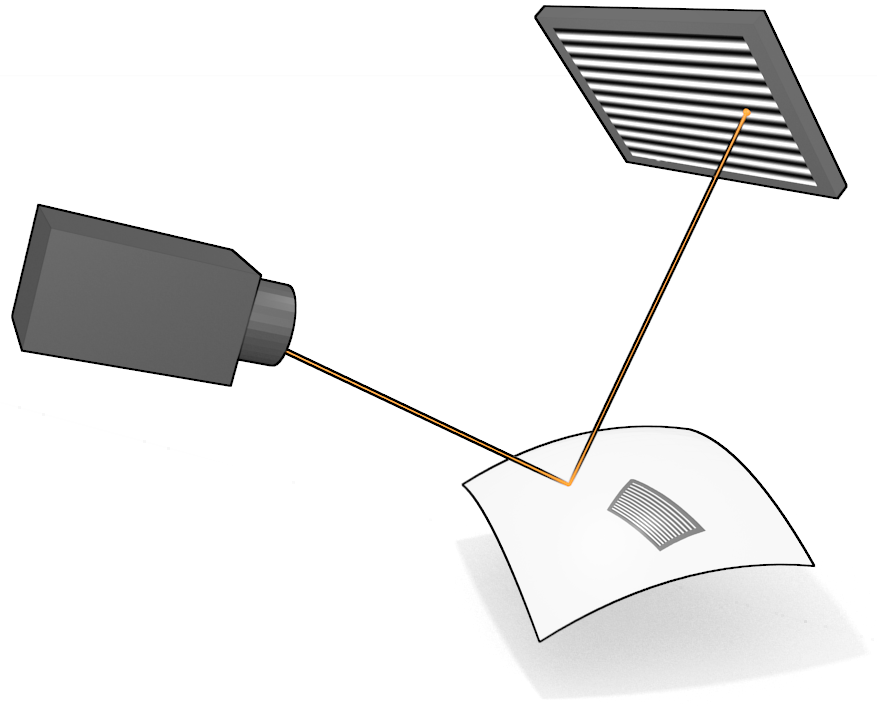}}\hfill
\subfigure[]{\label{fig:our_setup}\includegraphics[width=0.49\columnwidth]{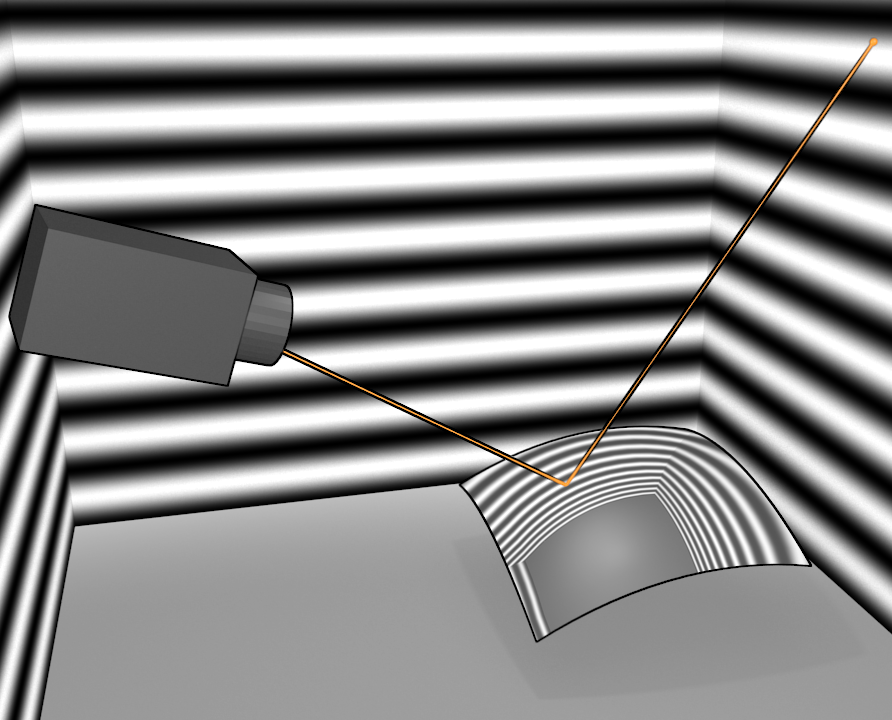}}\\
\subfigure[]{\label{fig:cave}\includegraphics[height=4cm]{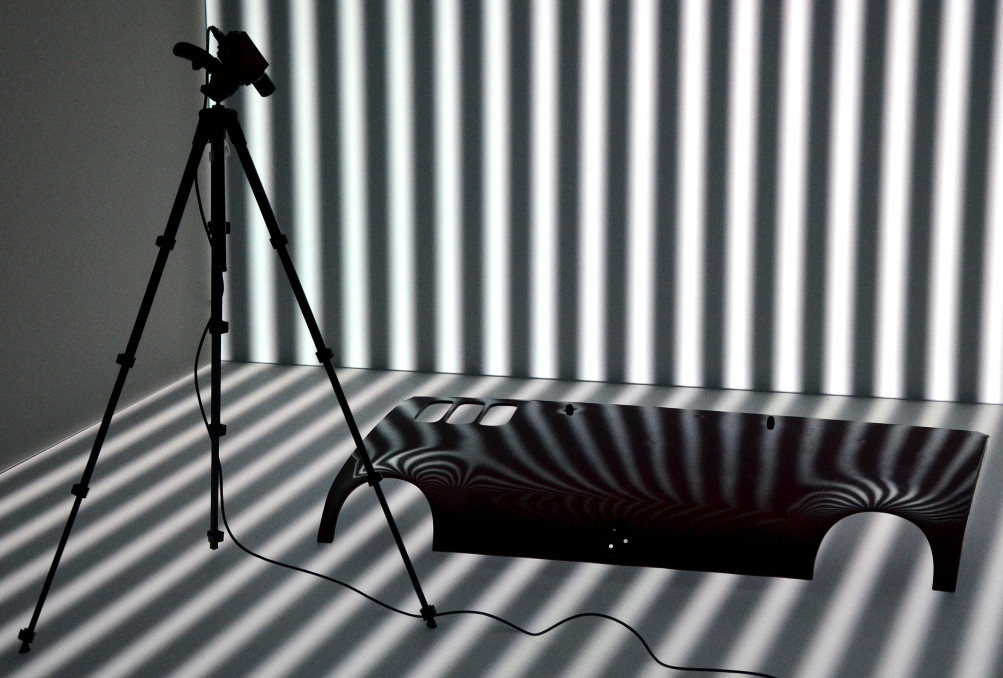}}\hfill
\subfigure[]{\label{fig:microcave}\includegraphics[height=4cm]{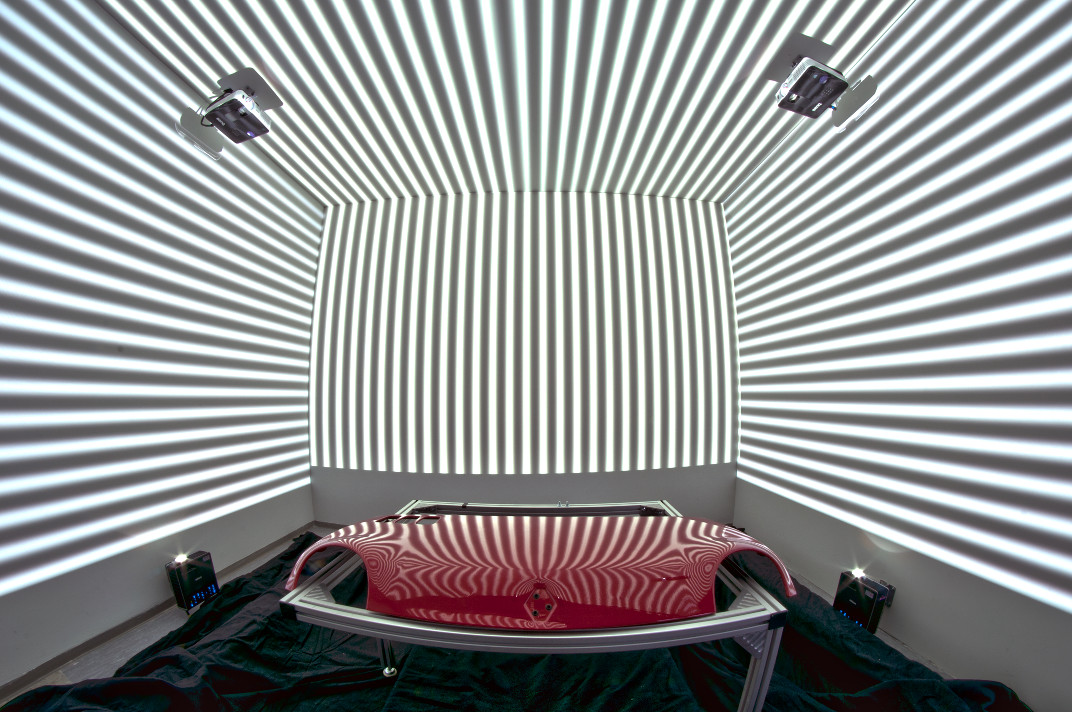}}
\caption{\subref{fig:classical_setup} Current deflectometric measurement setup consisting of a camera and a computer monitor, in between the unknown specular object. \subref{fig:our_setup} Our setup: we place the object in a CAVE which maximizes the encodable scene area \emph{and offers enough space to record images from any desired vantage point}. \subref{fig:cave} Cavlectometry recovers the shape of entire car parts like this engine hood from a single view. \subref{fig:microcave} The principles presented in this paper transfer without modifications to facilities orders of magnitude cheaper to realize than a CAVE.}
\label{fig:photosetup}
\end{figure}

Here, we propose not only to extend the illuminant to comprise the entire scene, but to do this in a way that supports acquisition of objects more than a meter in length from multiple vantage points. More concretely, we prove the feasibility of performing deflectometry in an \emph{Automatic Virtual Environment} (CAVE), which we dub \emph{cavlectometry}. A CAVE is a room-sized cube consisting of up to six back-projected walls, see Figs.~\ref{fig:our_setup} and~\subref{fig:cave}. In its default opera\-ting mode, users stand inside the CAVE and, while wearing stereo glasses and tracking devices, experience virtual 3-d scenes~\cite{Cruz-Neira1992}. ``Abusing'' it as an illuminant for deflectometry sparks a series of innovations: Rather than treating each of the walls separately, the encoding scheme introduced in Sect.~\ref{subsec:lightmap} -- our first contribution -- takes into account the sphere-like topology of the CAVE. In Sect.~\ref{sec:calibration}, we develop an calibration method which requires no additional data besides that used in later reconstruction, at the heart of which lies an algorithm for separating image regions arising from either Lambertian or specular surfaces. This is our second contribution. Cavlectometry allows us to reconstruct surface patches of unprecedented size from a single vantage point as demonstrated at hand of synthetic and real data sets (Sect.~\ref{sec:results}). Public deflectometry benchmarks are scarce with~\cite{Balzer2011} being the only notable exception. Thus, the final contribution we offer is to distribute all of our data, including raw images and decoded light maps, upon completion of the anonymous review process. Before we explain our method in detail, let us next put it in the context of the existing literature on specular surface reconstruction.

\subsection{Other related work}\label{subsec:relatedwork}

Normals of surfaces with hybrid reflection properties can be recovered from purely radiometric considerations~\cite{Ikeuchi1981,Healy1988,Schultz1994}. Not before correspondences are available between pixels and the scene points they portray, tools from geometrical optics can be leveraged: identifying a single point light source renders the computation of the light map trivial; then, under known camera motion, an initial point can be expanded into a surface curve by tracking the location where the highlight first appeared~\cite{Zissermann1989}. The standard structure-from-motion pipeline can also be enriched as to explicitly take into account specular reflections of a discrete set of scene points~\cite{Solem2004}. Zheng and Murata~\cite{Zheng2000} progress from isolated point sources towards a one-dimensional concentric illuminant. The accuracy of their approach however remains limited as long as some points on the light source remain indistinguishable. Circular reflection lines suffice nevertheless for special applications such as measuring the cornea in human eyes~\cite{Halstead1996} or surface interrogation~\cite{Ding2008,Sturm2009}. Savarese and Perona~\cite{Savarese2005} study reflections of a two-dimensional checkerboard pattern. 

Controllable illuminants were introduced to boost reliability and density of correspondences: The very first active deflectometry setup -- due to  Sanderson et al.~\cite{Sanderson1988} -- contained an array of light-emitting diodes (LEDs) which could be switched on sequentially to detect them in the camera image. The acquisition time can be reduced by showing binary codes in parallel~\cite{Nayar1990}. Both papers address the ambiguity by assuming quasi-parallel illumination. A large body of literature on the subject exists in the field of optical metrology starting with~\cite{Perard1997} that suggests improvements if the LED array is replaced by a commodity computer monitor. Knauer et al.~\cite{Knauer2004} describe the theoretic limits of the method in\-vol\-ving phase-shifted sine patterns as codes. Light map measurements can be further enhanced by color displays~\cite{Tarini2005}. An active variant of Savarese's method is presented in~\cite{Rozenfeld2011}.

Gauge ambiguities can be eliminated by integrability considerations~\cite{Hicks2005,Liu2013} or \emph{specular stereo} ~\cite{Wang1993,Donner2004,Bonfort2006,Weinmann2013}. In the special case of infinitesimal displacement of any participant of the imaging chain -- camera, surface, or scene -- the deflectometric image is perturbed by a \emph{specular flow} field. Blake treats the case of a moving \emph{observer} in a series of papers culminating in~\cite{Blake1991}, which Waldon and Dyer~\cite{Waldon1993} extend by adding photometric effects and interreflections to the model. Like in~\cite{Zissermann1989}, the trajectories of virtual features in the image plane give rise to reconstructions of one-dimensional subsets of the sought-after surface~\cite{Oren1995}. They are linked with the caustic curves known in geometric optics~\cite{Swaminathan2002}. Roth's and Black's~\cite{Roth2006} method delivers dense two-dimensional reconstructions from specular flow ge\-ne\-ra\-ted by a known camera motion. The authors utilize theoretical results from~\cite{Chen2000} where, with the objective of speeding up ray tracing along piecewise linear paths, it is shown how the points of deflection, i.e., the vertices of the path, behave w.r.t. to changes of the ray emitter. Shape cues may be recovered from the flow field alone, i.e., without image-scene correspondences, when the \emph{scene} is moving relative to object and camera~\cite{Adato2010}. A closed-form model of specular flow on arbitrarily deforming \emph{surfaces} is derived in~\cite{Lellmann2008}. Estimation of specular flow hinges only on the natural scene and can thus handle surfaces of arbitrary size and shape complexity but, as a downside, does not permit metrically precise reconstruction. A comprehensive summary of the state of the art in deflectometry can be found in several recent survey papers~\cite{Ihrke2010,Balzer2010,Reshetouski2013}. 
\section{Data acquisition}
\subsection{Measurement of the light map}\label{subsec:lightmap}
\begin{figure}[tb]
\centering
\def\svgwidth{0.8\columnwidth}
{\normalsize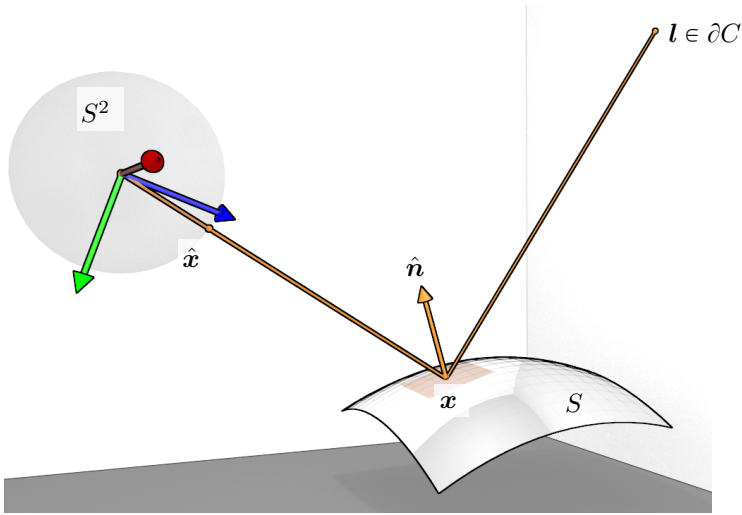}
\caption{Geometry of reflection. The image of the \emph{light map} is the point $\bm{l}$ that is seen via a point $\bm{x}$ on the specular surface $S$ under the direction~$\hat{\bm{x}}\in S^2$.}
\label{fig:geometry}
\end{figure}

In the following, matrices are in bold, vectors bold italic. Vectors $\bm{x}\in\mathbb{R}^3$ of unit length are marked by a hat, i.e., $\|\hat{\bm{x}}\|=1$. Each pixel in the image plane of a calibrated ca\-me\-ra corresponds to a direction $\hat{\bm{x}}$, under which the associated sight ray leaves the projection center. The relationship between the two is determined by the intrinsic parameters of the camera. Where the ray intersects the surface in $\bm{x}$, it gets ``deflected'' according to the geometric law of reflection and travels on until it hits a scene point $\bm{l}$. For the purposes of geometric modeling, the CAVE is a cube $C$ with boundary $\partial C$ and side length $2h=3048$ $\mathrm{mm}$. If we neglect multiple reflections and occlusions induced by the camera, the scene point comes to lie on one of the faces of $C$. Apart from the object under inspection, the scene \emph{is} $\partial C$ and vice-versa. Altogether we obtain a mapping $l:S^2\to \partial C$, which assigns to a viewing direction $\hat{\bm{x}}$ the point $\bm{l}$ seen by the pixel associated with that direction, see Fig.~\ref{fig:geometry}. This is precisely the light map defined in Sect.~\ref{subsec:motivation}. Its measurement must be the first step of any deflectometric method.
\subsubsection{Phase shifting}
Although optical phase-shift codes originate from a variety of established applications like optical interferometry, synthetic aperture radar, and magnetic resonance imaging, its features make it the ideal candidate for estimation of the light map: Gradually phase-shifted sine patterns facilitate subpixel accuracy and are relatively insusceptible to disturbances of absolute intensity values, e.g., caused by staining of the specular surface. As in a typical setup (Fig.~\ref{fig:photosetup}) the camera is focused onto the surface under inspection, limited depth of field exposes the imaged code patterns to blurring. A sine pattern, however, can pass the optical system -- which in this case resembles a low-pass filter -- unaffected as long its spatial frequency is sufficiently low. In fact, a smoothing of quantization artifacts of the monitor or projector is even advantageous.

\begin{figure}[t]
\centering
\subfigure[Naive approach]{\label{fig:cornerold}
\includegraphics[width=0.45\columnwidth]{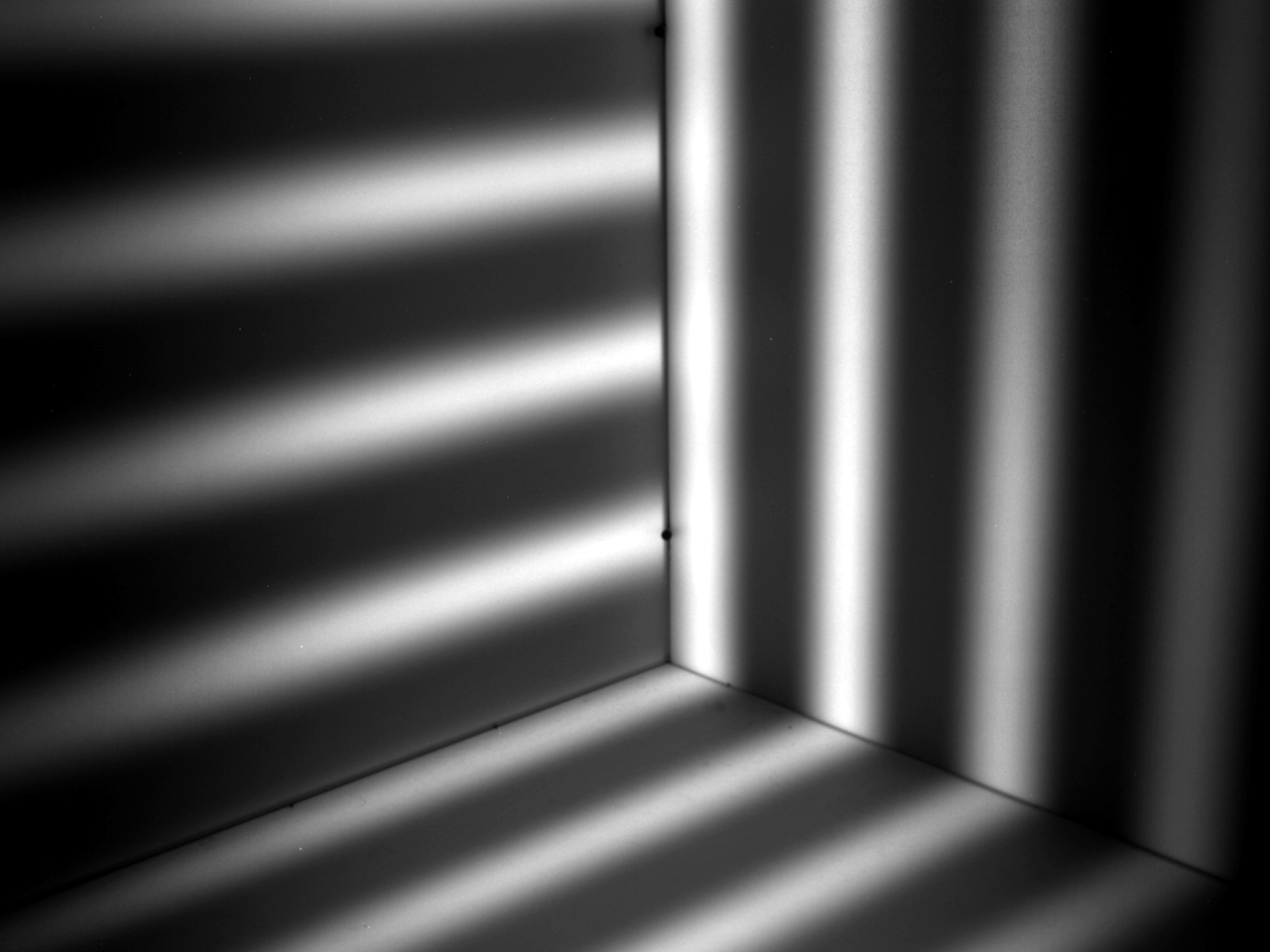}\hspace{0.1cm}
\includegraphics[width=0.45\columnwidth]{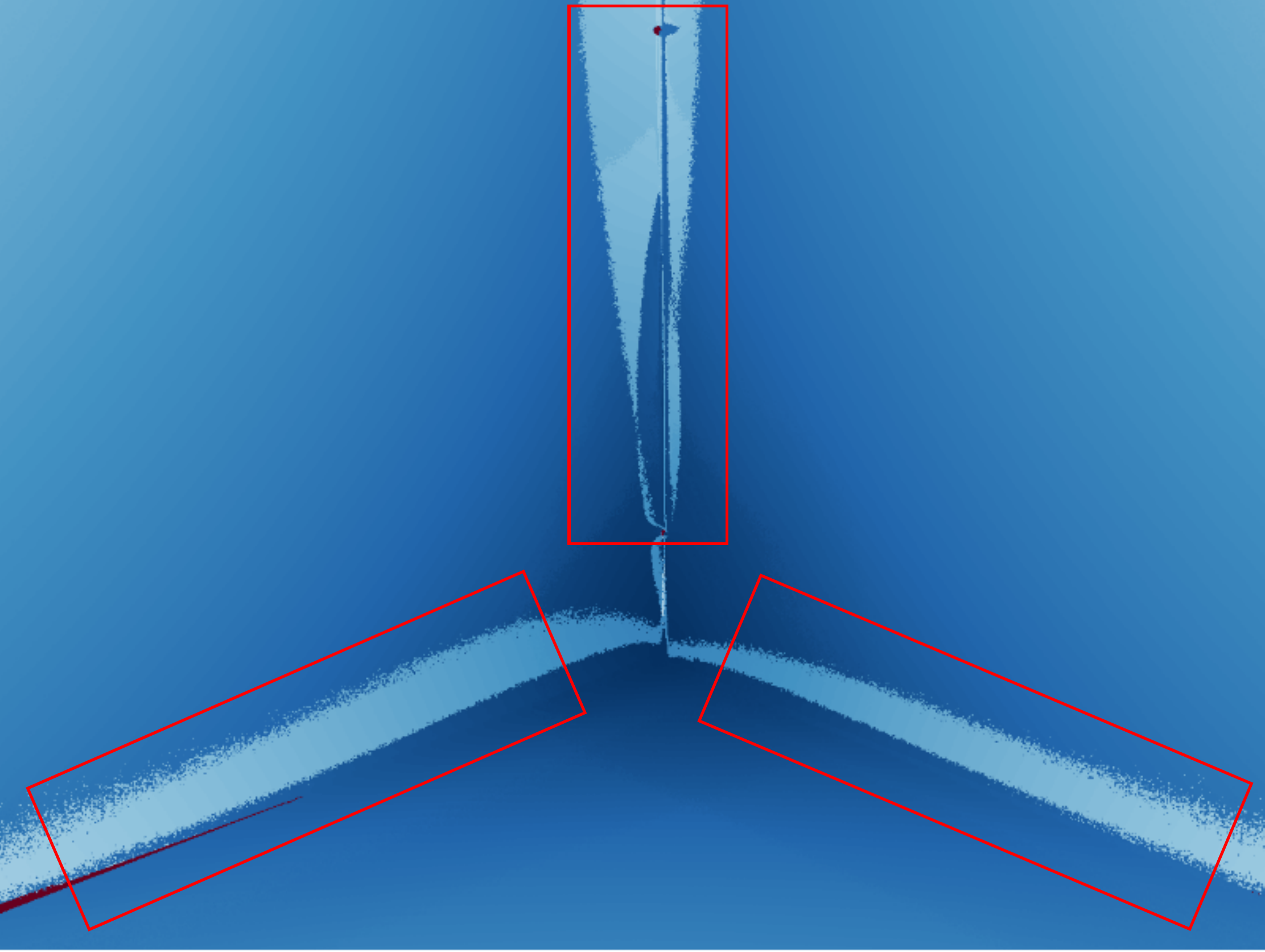}}\\
\subfigure[Proposed approach]{\label{fig:cornernew}\includegraphics[width=0.45\columnwidth]{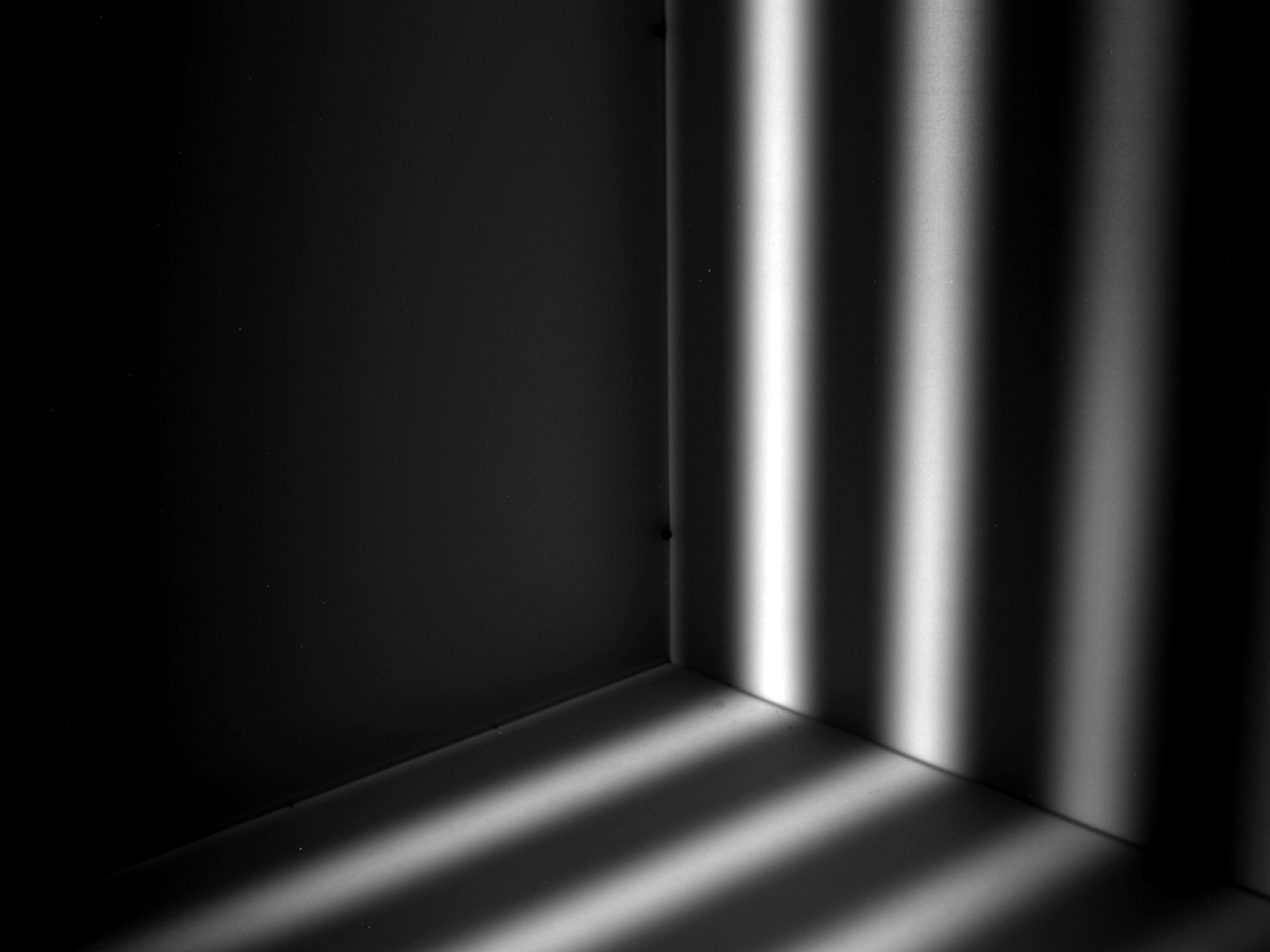}\hspace{0.1cm}
\includegraphics[width=0.45\columnwidth]{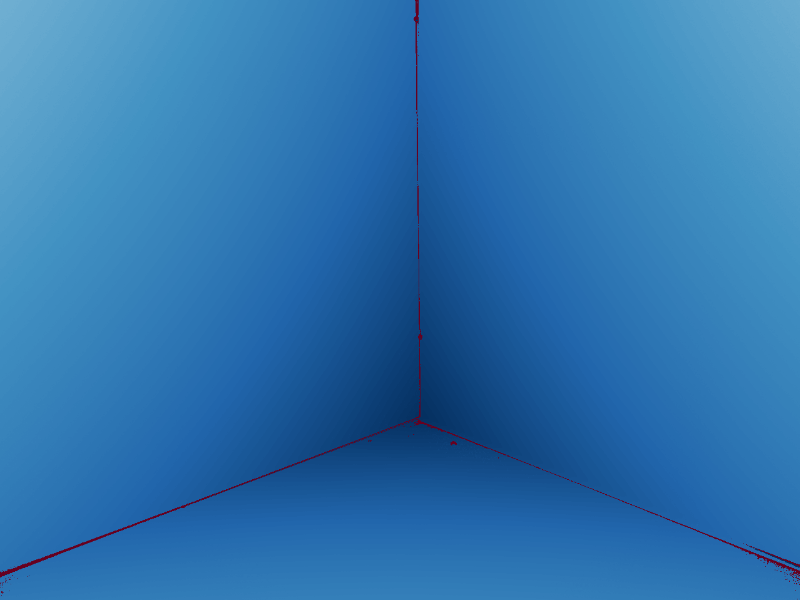}}
\caption{Benefit of a proper encoding strategy: \subref{fig:cornerold} Different walls of the CAVE illuminate each other causing \emph{leakage} of the light map close to the transition between two walls (red boxes). \subref{fig:cornernew} This effect can be mitigated by using only patterns which are symmetric w.r.t. all edges parallel to the shift direction.}
\label{fig:cornerdecode}
\end{figure}

Three phase shifts are sufficient to recover a single spatial coordinate uniquely; additional ones may compensate for non-uniformities (e.g., from gamma nonlinearity, etc.). Owing to the fact that phase information is uniquely defined only within the range of one wavelength of the pattern, decoding must be complemented by a so-called \emph{phase unwrapping}. Explaining the codification methodology in detail is beyond the scope of this paper; we must refer the interested reader to~\cite{Ghiglia1998} among others. We do emphasize here that care must be taken when displaying phase-shifted sine patterns on the walls of the CAVE simultaneously, an issue that never arose in previous systems which were all designed to run on disjoint rectangular screens. The naive approach would be unfolding the cube into the canonical net for cube mapping and shift the patterns along the axes of some local coordinate system. Experimental evidence, however, suggests that treating each face regardless of the others causes problems at the transitions between two maps, see Fig.~\ref{fig:cornerdecode}. Here, due to self-illumination, a pattern leaks from one wall into the other. Our key insight is that this is uncritical so long two patterns are symmetric across an edge of $\partial C$, i.e., if they do not meet orthogonally, at different frequencies or phases. An encoding scheme which fulfills this requirement by moving patterns only parallel to the edges of $\partial C$ is shown in Fig.~\ref{fig:phasecave}. 

\subsubsection{Face encoding}
\begin{figure}[b]
\centering
\subfigure[$x$]{\label{fig:cs}%
\begin{tabular}[b]{c}
\includegraphics[width=0.3\columnwidth]{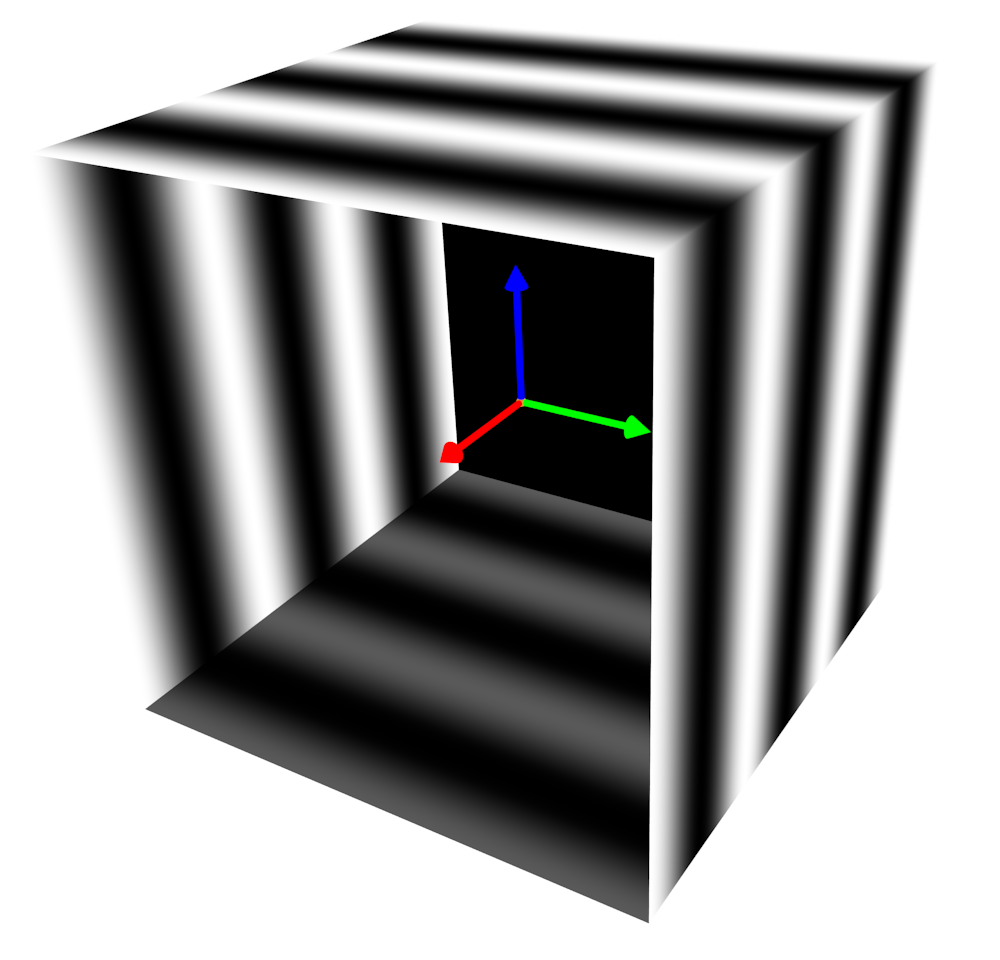}\\
\includegraphics[width=0.28\columnwidth]{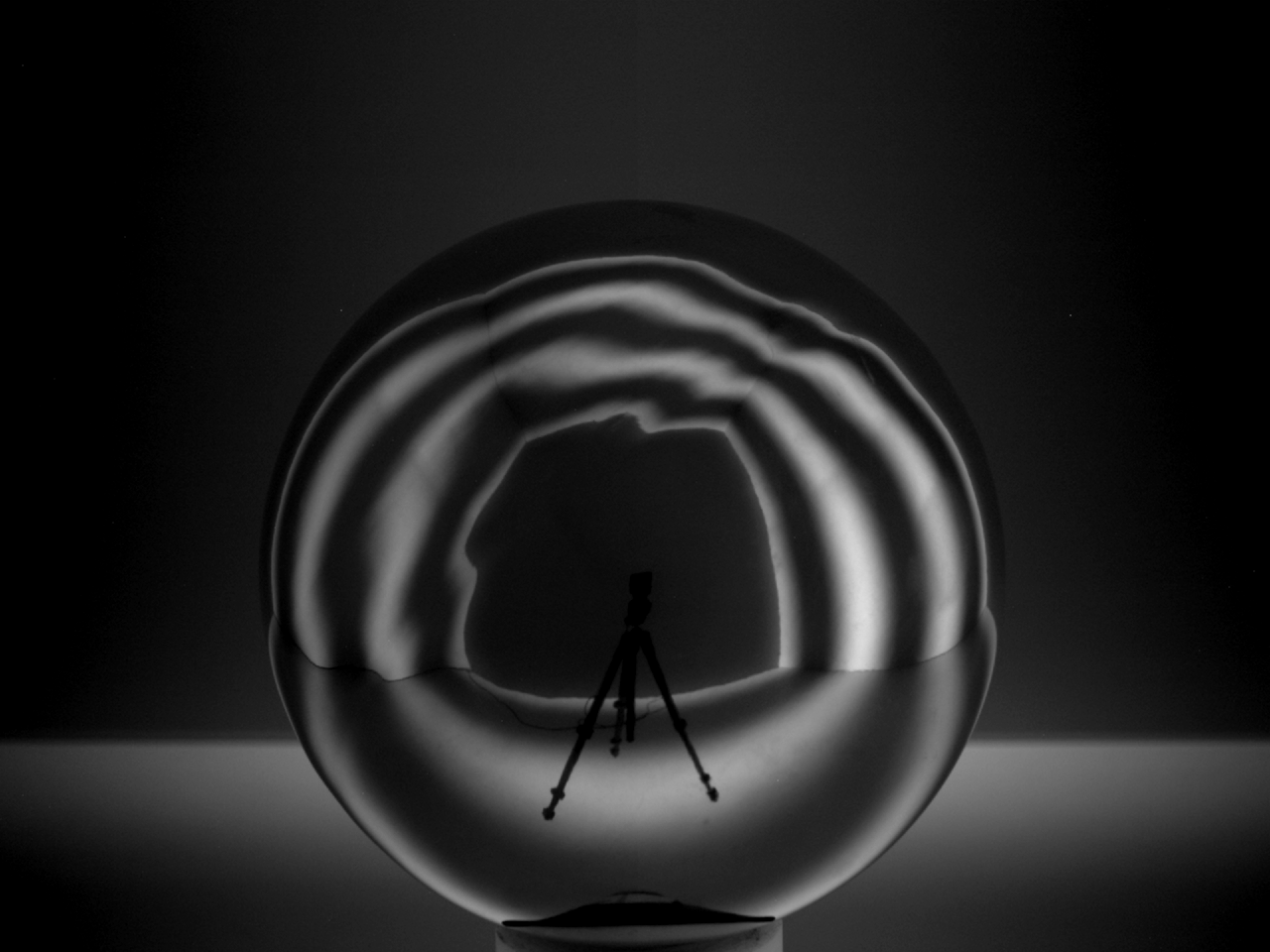}\hfill
\end{tabular}}%
\subfigure[$y$]{%
\begin{tabular}[b]{c}
\includegraphics[width=0.3\columnwidth]{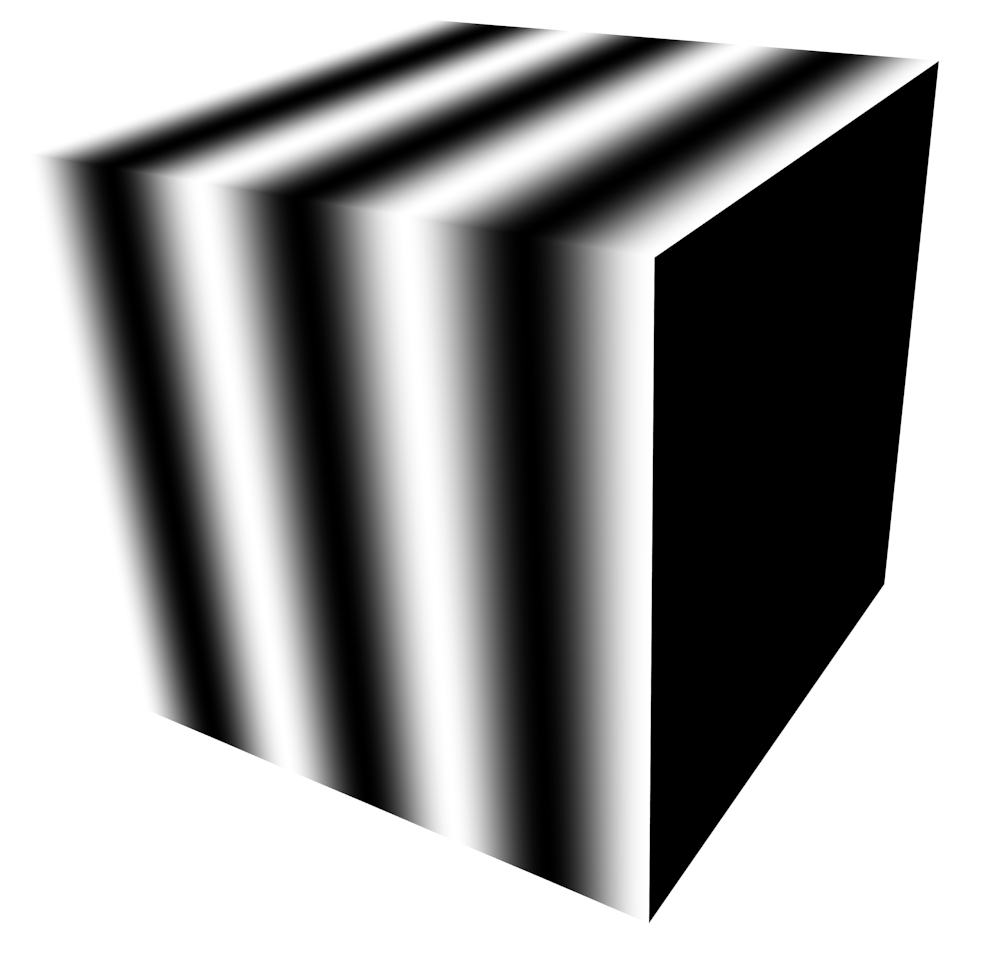}\\
\includegraphics[width=0.28\columnwidth]{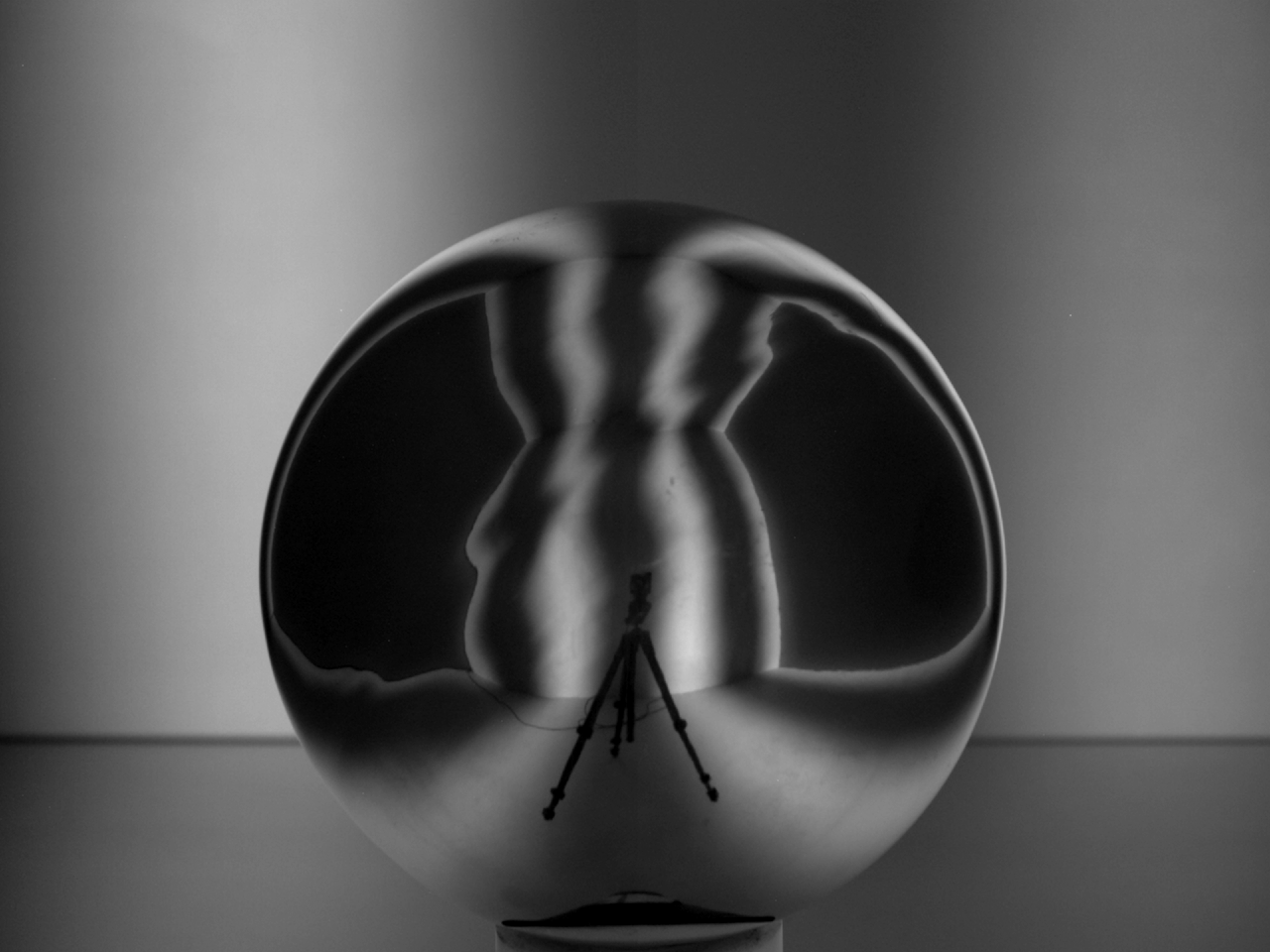}\hfill
\end{tabular}}%
\subfigure[$z$]{%
\begin{tabular}[b]{c}
\includegraphics[width=0.3\columnwidth]{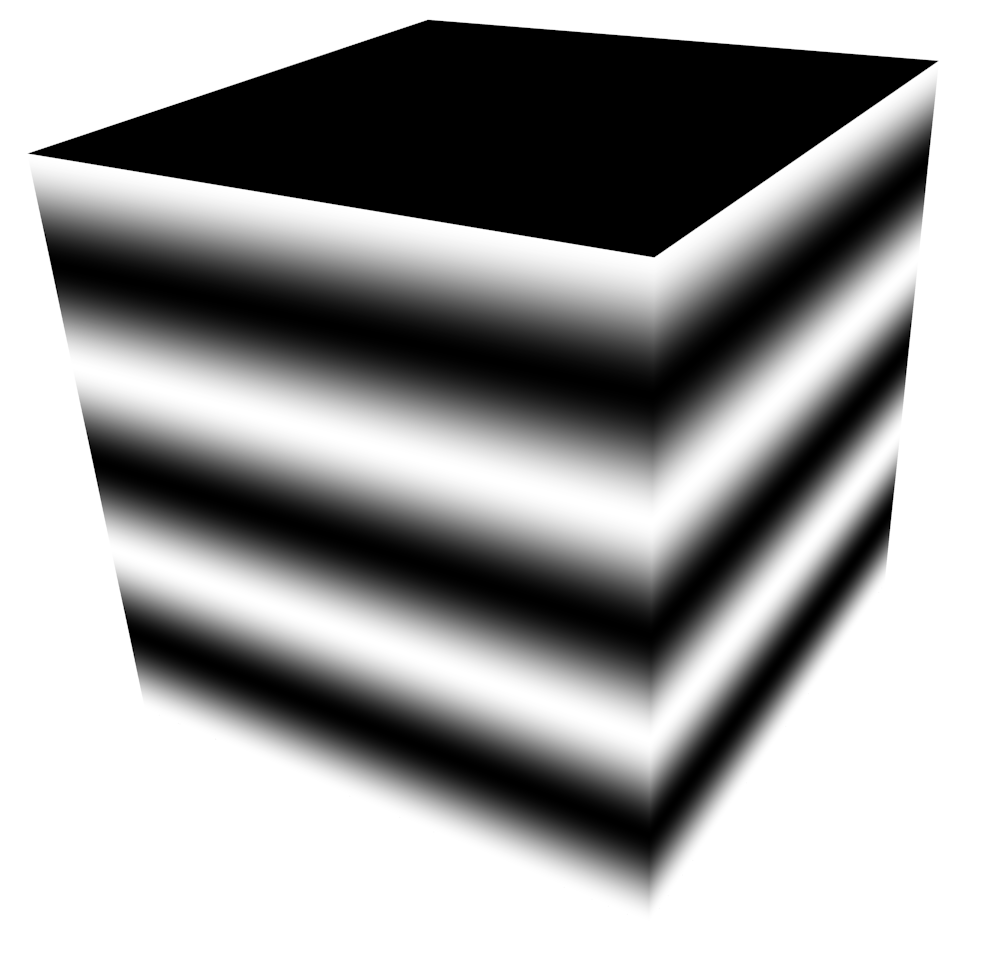}\\
\includegraphics[width=0.28\columnwidth]{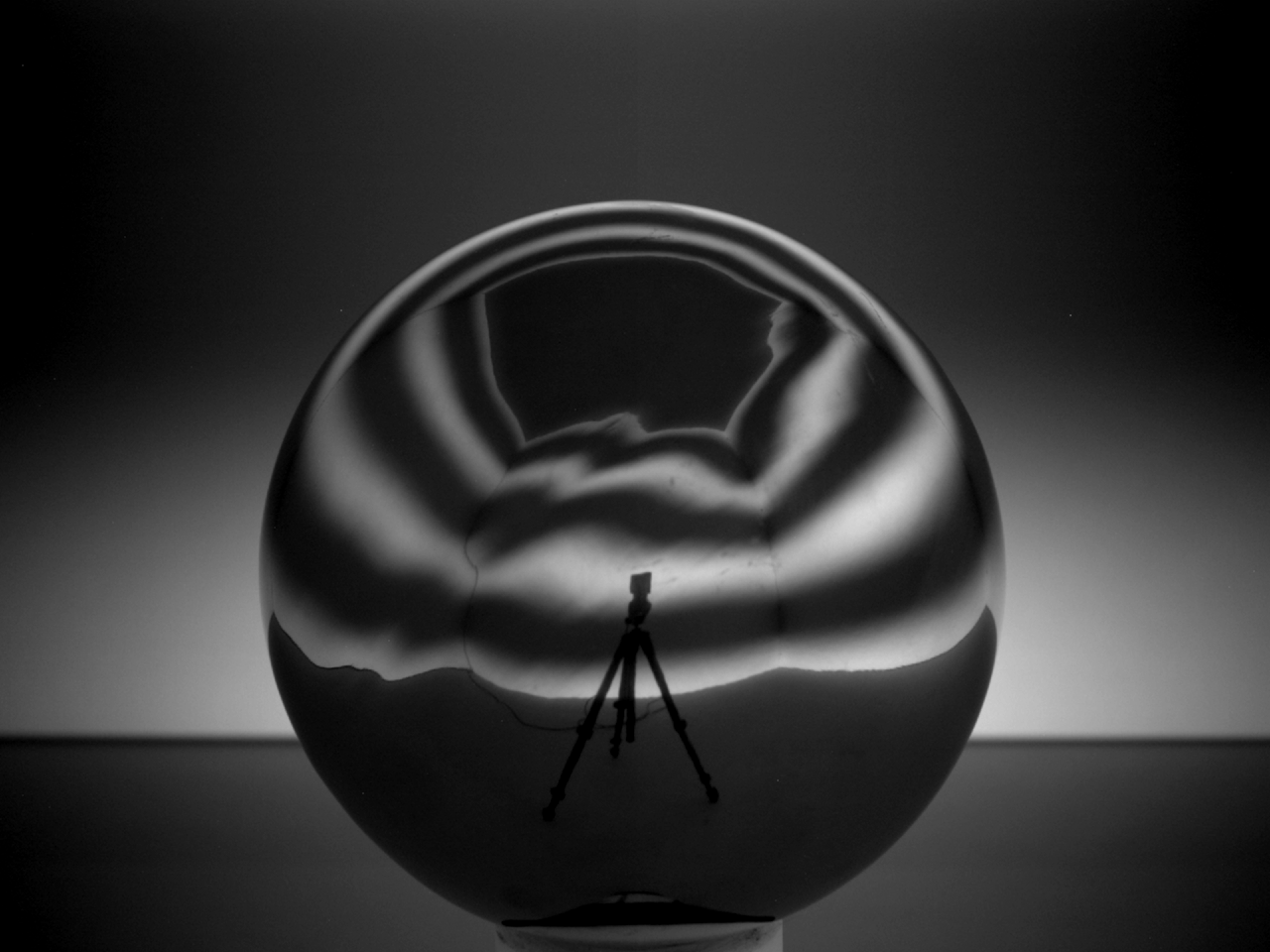}
\end{tabular}}%
\caption{Arrangement of sine patterns shown on the CAVE walls along three spatial directions. During measurement, these are displayed at multiple phases and frequencies.}
\label{fig:phasecave}
\end{figure}

The inherent structure of this scheme is easily verified: each subsequence encodes one of the three global Cartesian coordinates. Only a single $\mathrm{bit}$ of information is missing, prompting a second coding stage. To illustrate the problem, suppose we are looking for the location of a point $\bm{l}$ on one of the CAVE walls, say $\bm{l}=(l_x,l_y,l_z)\T$ where $l_x=h$. Applying the sequences in~Fig.~\ref{fig:phasecave} will provide the values of $l_y$ and $l_z$ there. The subsequence shifting in $x$-direction will leave $\bm{l}$ dark. Putting these facts together, we conclude that $\bm{l}$ will be on a plane $x=\mathrm{const.}$ but we do not know whether it is the one defined by the equality $x=h$ or $x=-h$. The sign of $h$ is exactly the missing bit. A single image should suffice to set it correctly. The problem of self-illumination, however, persists. In fact, the leakage of a pattern shifting in one of the three coordinate directions into the two or\-tho\-go\-nal faces that remain dark meanwhile is even more severe. For this reason, we resort to a complete detection of each of the faces of $C$. There are altogether six of them prompting the use of a binary code of length $3$ $\mathrm{bit}$. An illuminated wall represents a bit value of $1$, see Fig.~\ref{fig:walls}. Robustness is further improved with the help of a \emph{differential code}: Thereby, a bit value of $1$ is encoded by the \emph{transition} from white to black. Conversely, a bit value of $0$ corresponds to an on-switching of the illumination. The differential face coding amounts to in total six images per light map measurement.
\begin{figure}[t]
\centering
\subfigure[Face ID]{\label{fig:walls}\includegraphics[width=0.45\columnwidth]{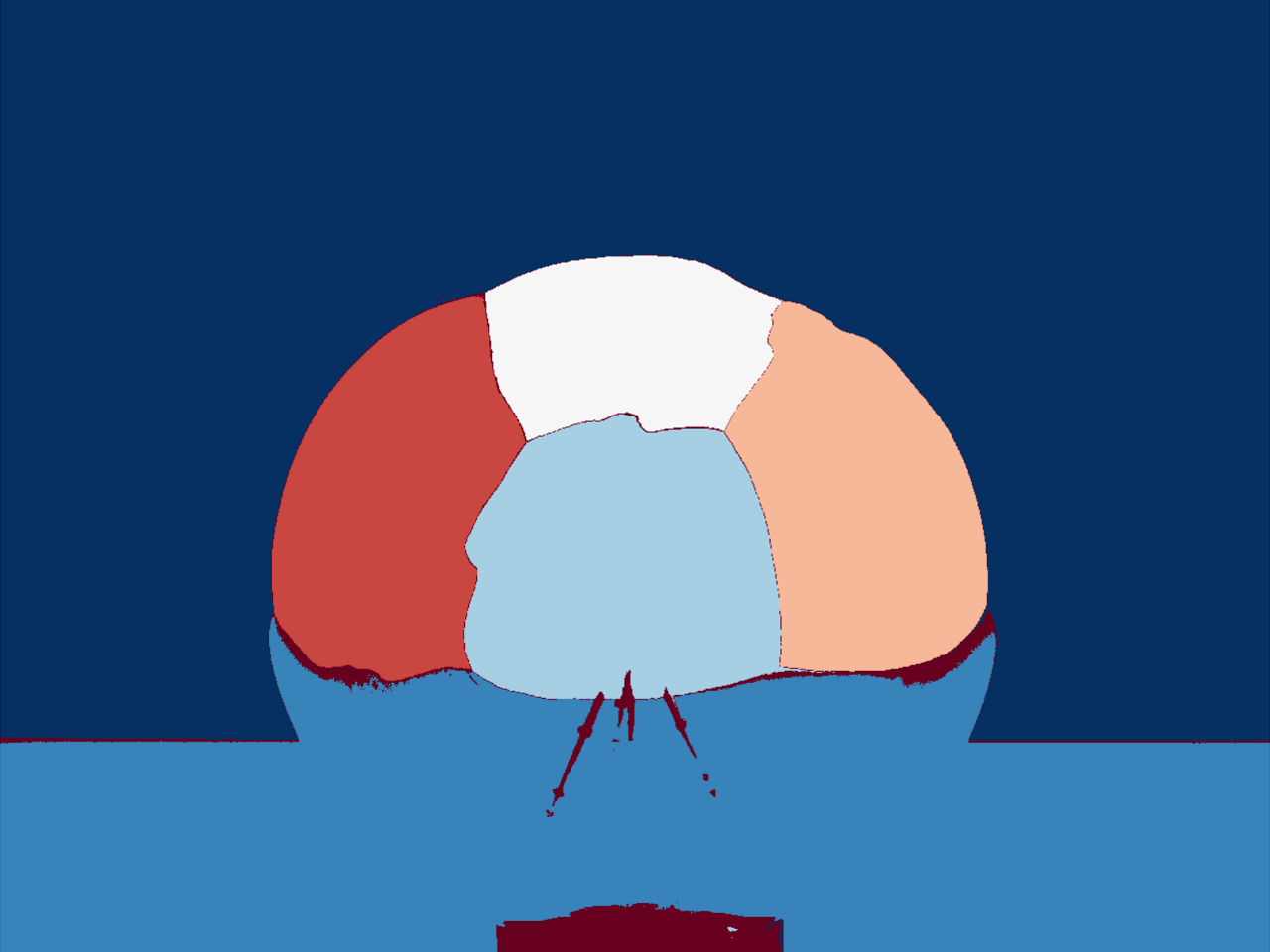}}\hspace{0.1cm}
\subfigure[Full lightmap (view in color)]{\label{fig:l}\includegraphics[width=0.45\columnwidth]{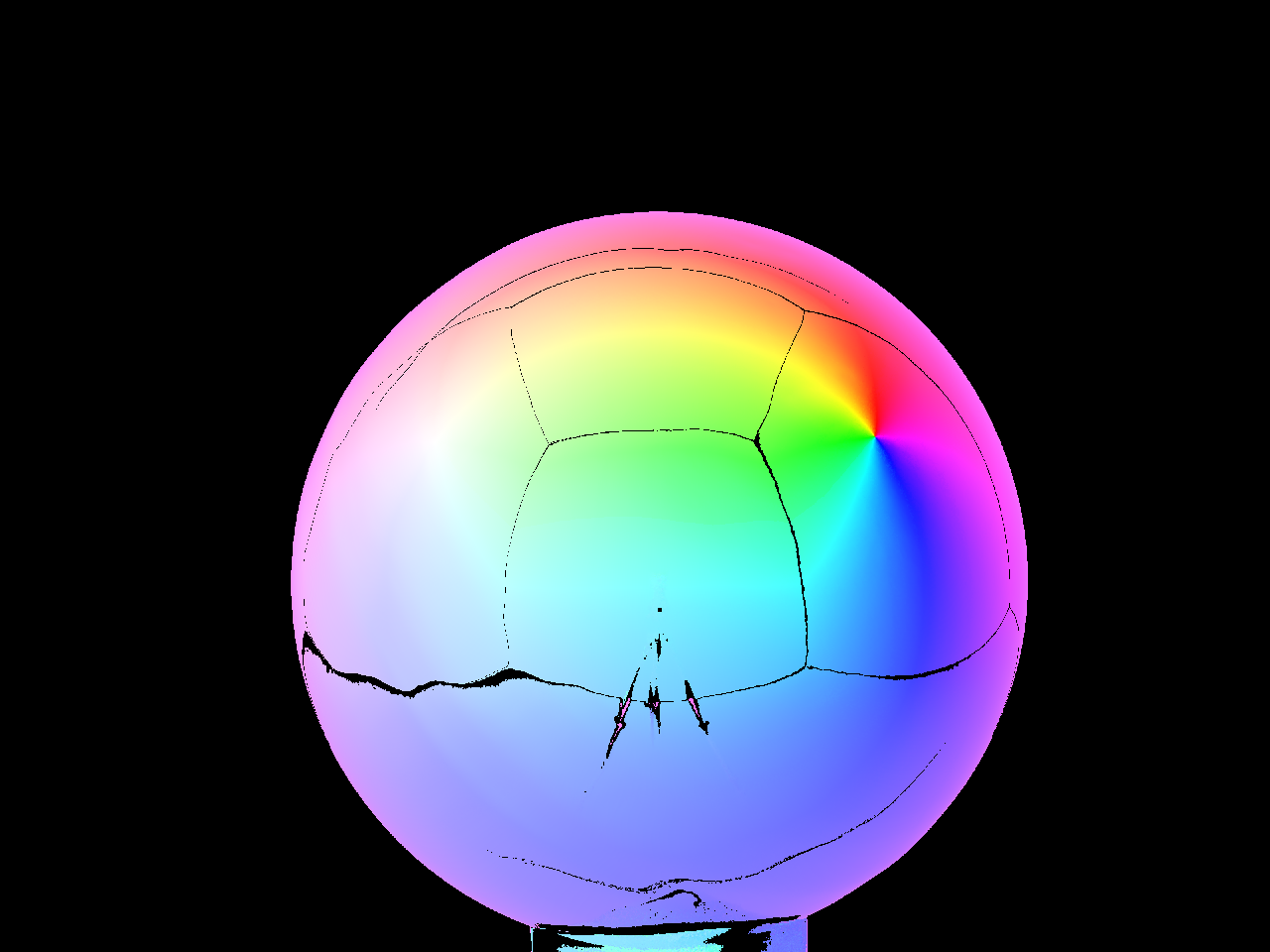}}
\caption{\subref{fig:walls} A binary code identifies each separate wall uniquely. \subref{fig:l} Visualization of the full vector-valued light map $l$. The color map assigns an HSV coordinate to each $\hat{\bm{l}}\in S^2$ with hue and saturation varying proportionally to azimuth respectively elevation while the value is kept constant at $1.0$.}
\label{fig:walls}
\end{figure}
\section{Calibration}\label{sec:calibration}
In order to avoid screen occlusions, we advocate re\-cor\-ding multiview datasets in a monocular fashion, whereby measurements are taken sequentially while the camera is displaced between two captures. To convert a measured correspondence between $\hat{\bm{x}}$ and $\bm{l}=l(\bm{x})$ into a reasonable normal estimate, both vectors have to be represented in a common reference frame. On one hand, $\hat{\bm{x}}$ is usually given w.r.t. the frame that is attached to the camera and moves along with it to each new vantage point. On the other hand, $\bm{l}$ is naturally expressed in terms of the frame at the barycenter of the CAVE (Fig.~\ref{fig:cs} top), which remains static and is thus selected as our world coordinate system. The transformation between the two is an element $g=(\mathbf{R},\bm{t})$, $\mathbf{R}\in\SO(3)$, $\bm{t}\in\mathbb{R}^3$, of the rigid-motion group $\SE(3)$, which -- as initially unknown -- needs to be determined through calibration. 

It turns out that $g$ comes as a byproduct of the very same light map $l$ used for reconstruction later on (Sect.~\ref{sec:reconstruction}). The basic idea of the following procedure is to identify those regions that are imaged \emph{without} intermediate reflections and to localize the camera by means of the appendant scene-image correspondences:
\begin{enumerate}
\item Consider the pullback of the light map $l^*:D\to \partial C$ to the image plane $D\subset\mathbb{R}^2$. It can be calculated knowing the intrinsic parameters of the camera, which we initially estimate by a standard method~\cite{Zhang2000}. The extrinsic calibration process starts with a -- possibly very coarse -- segmentation of $l^*$, which builds upon the fact that walls seen directly generate piecewise homographies. One of the distinct properties of homographies is that they maintain cross-ratios, whose computation, however, may be numerically unstable. Instead, we propose a simpler -- less discriminative yet effective -- inference criterion based on colinearity. Denote by $\bm{l}_{ij}$ the value of the light map at pixel $\bm{p}:=(u_i,v_j)\in D$ in the discretized image plane. If $l^*$ behaves locally like a homography around $\bm{p}$, then the points $\bm{l}_{i,j-1},\bm{l}_{i,j},\bm{l}_{i,j+1}$ and $\bm{l}_{i-1,j},\bm{l}_{i,j},\bm{l}_{i+1,j}$ will be respectively colinear, in other words, $r^x_{ij}=r^y_{ij}=1$ where
\[
r_{ij}^x:=\rank (\bm{l}_{i-1,j},\bm{l}_{i,j},\bm{l}_{i+1,j})
\]
and
\[
r^y_{ij}:=\rank (\bm{l}_{i,j-1},\bm{l}_{i,j},\bm{l}_{i,j+1}).
\]
We can define a scoring function $s:D\to\mathbb{N}$ by
\begin{equation}\label{eq:scoringfunction}
s(u_i,v_j):=|1-r^x_{ij}r^y_{ij}|,
\end{equation}
which vanishes where both rank conditions are satisfied (Fig.~\ref{fig:sssega}). A first segmentation of $l$ respectively $l^*$ is obtained by binarizing $s$ with threshold $0.5$ (Fig.~\ref{fig:sssegb}). Following morphological enhancement, this segmentation is refined with the help of a watershed transform~\cite{Meyer1992}, see Fig.~\ref{fig:sssegc}.
\begin{figure}[t]
\centering
\subfigure[$s(u_i,v_j)$]{\label{fig:sssega}\includegraphics[width=0.45\columnwidth]{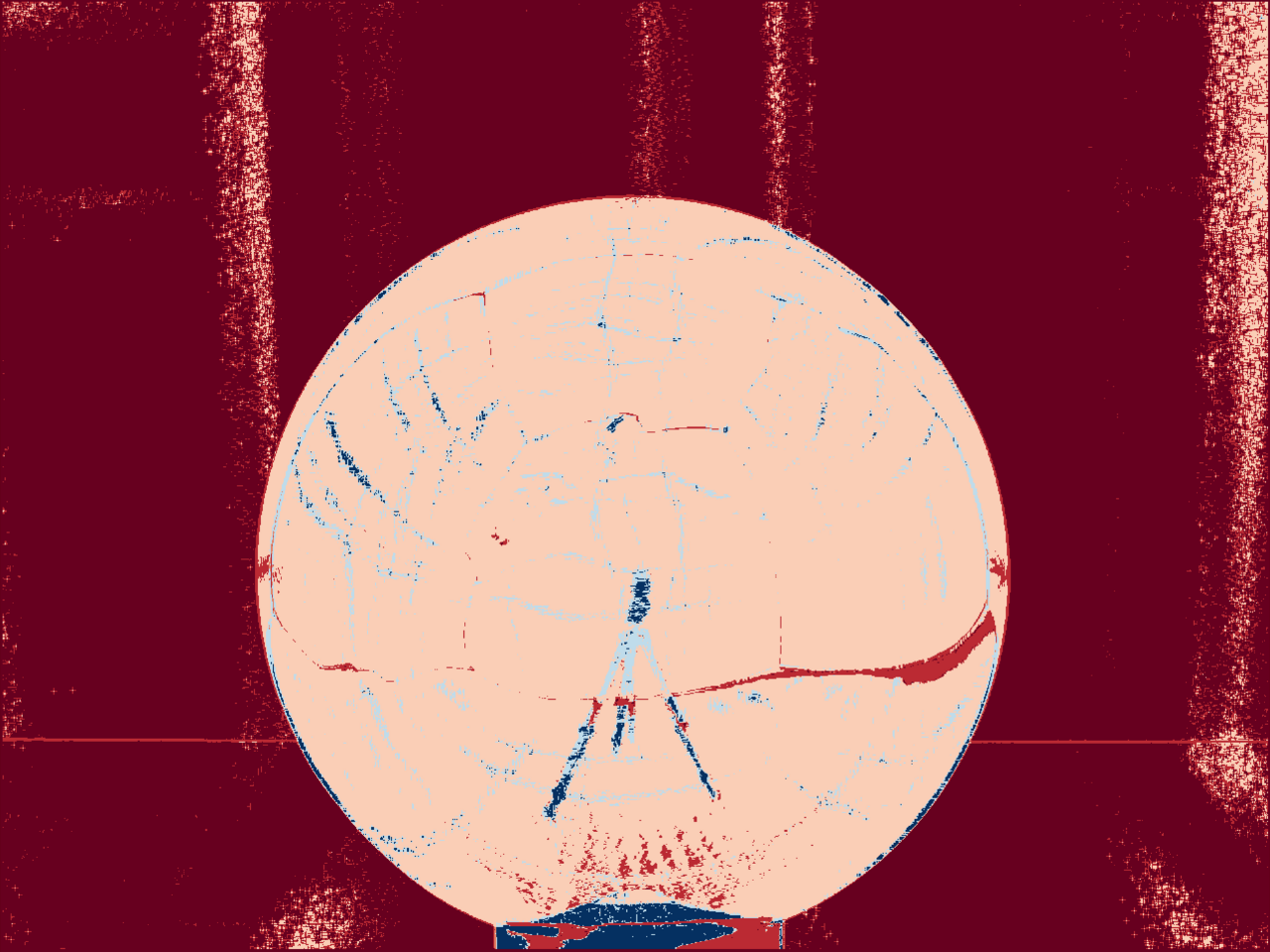}}\hspace{0.1cm}
\subfigure[]{\label{fig:sssegb}\includegraphics[width=0.45\columnwidth]{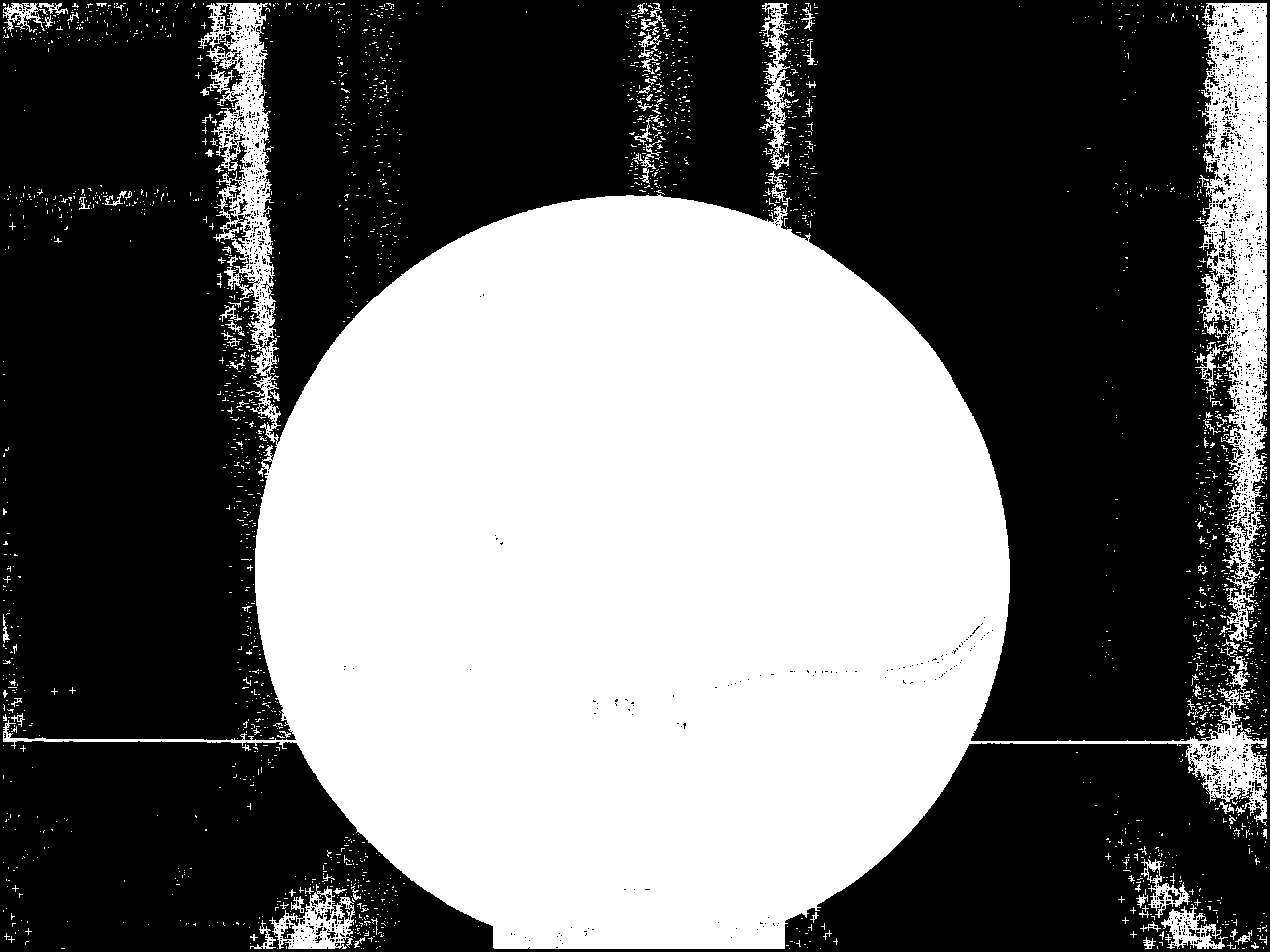}}\\
\subfigure[]{\label{fig:sssegc}\includegraphics[width=0.45\columnwidth]{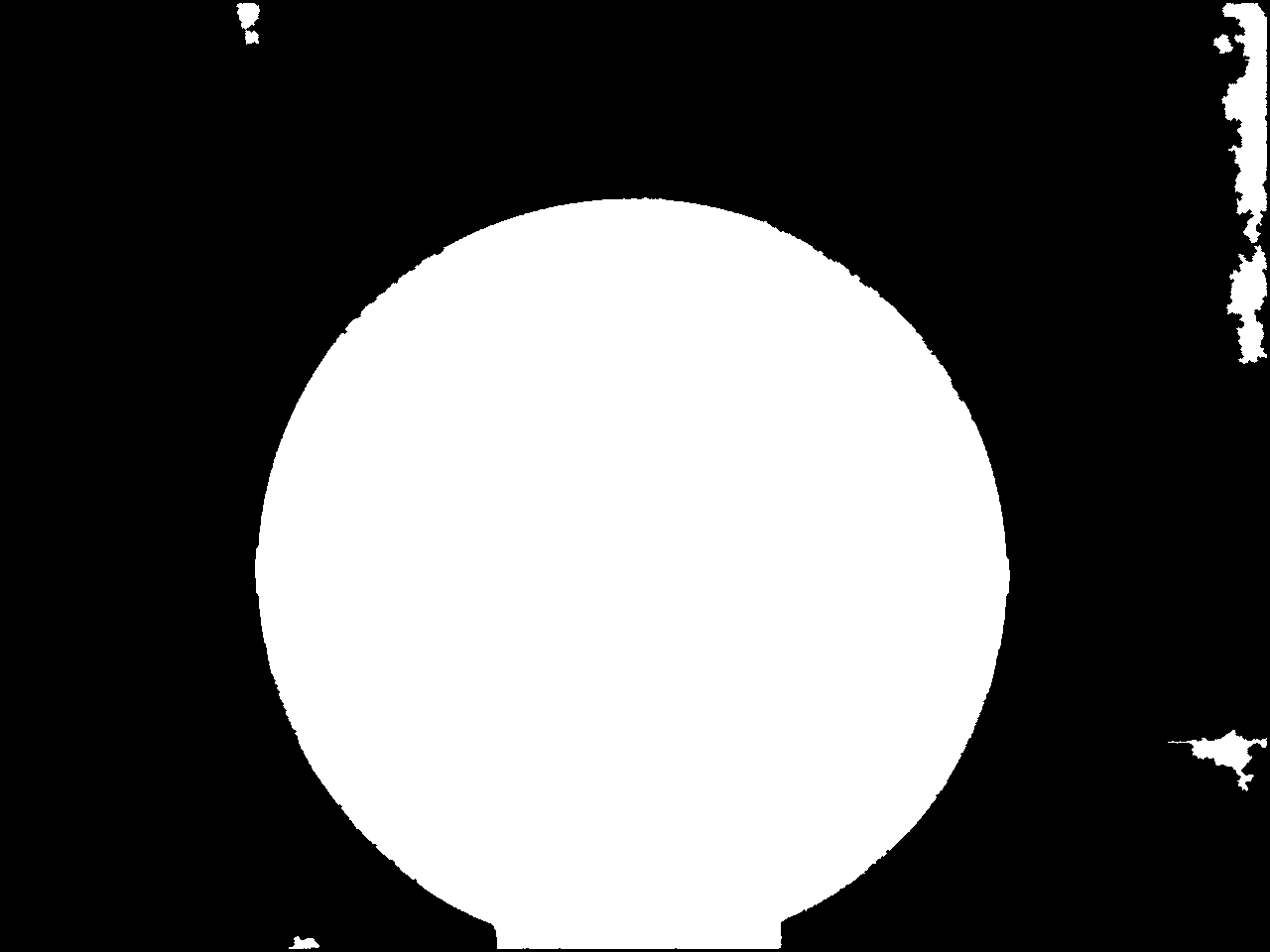}}\hspace{0.1cm}
\subfigure[$\chi_{\mathrm{b}}$]{\label{fig:sssegd}\includegraphics[width=0.45\columnwidth]{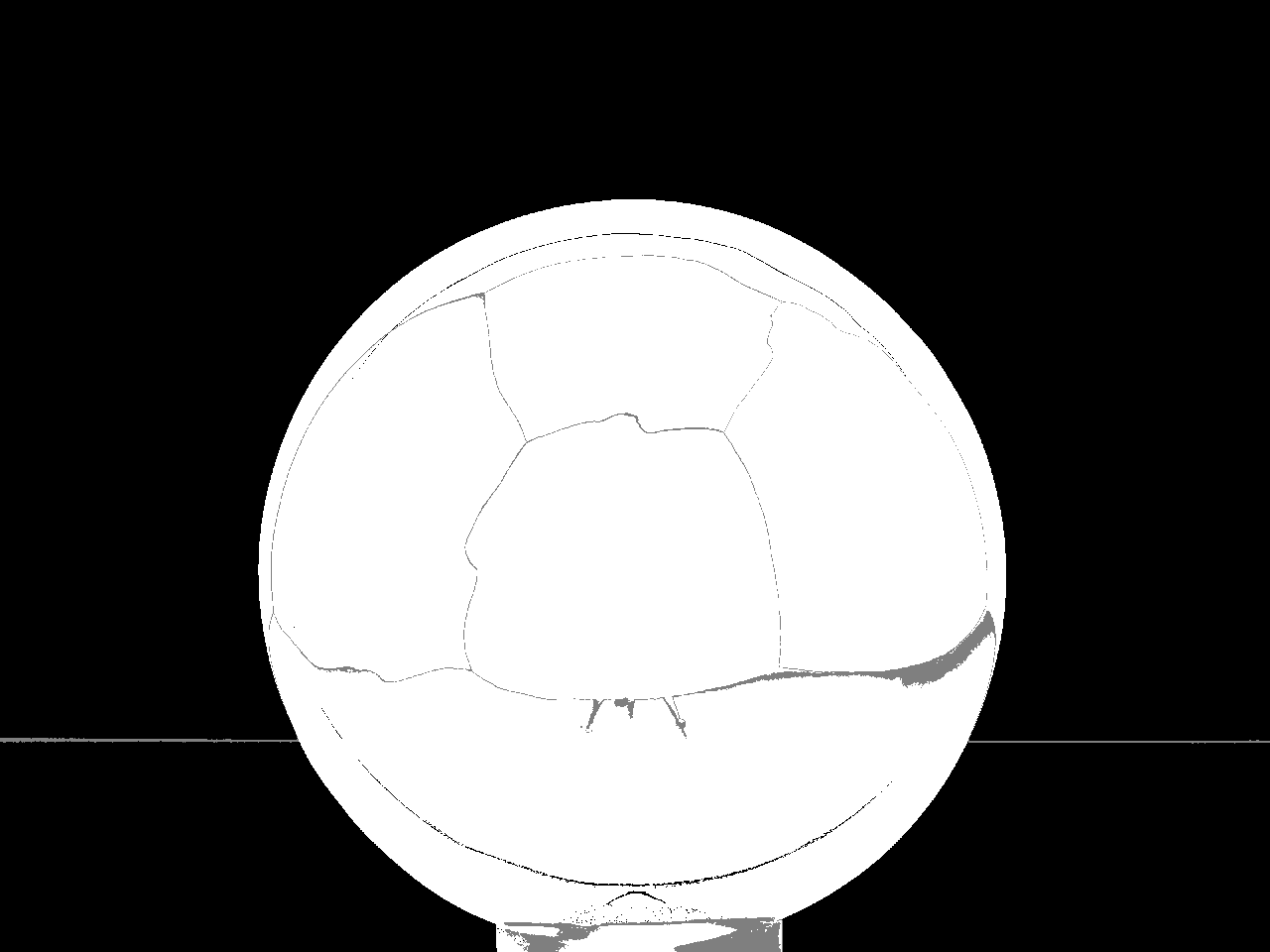}}
\caption{Segmentation of deflectometric images of strongly-specular objects: \subref{fig:sssega} Foreground score based on local ranks of the light map. \subref{fig:sssegb} Initial segmentation of \subref{fig:sssega} by thresholding. \subref{fig:sssegc} Enhancement by watershed transform. \subref{fig:sssegd} The final segmentation is obtained after extrinsic calibration by thresholding the backprojection error.}
\label{fig:ranks}
\end{figure}
\item Observe that there are degenerate cases in which the specular surface itself affords a homography, e.g., if it is locally planar or reflects into a single point (i.e., the light map is constant). We can, however, safely assume that the background occupies a significant amount of the image. Otherwise, we would hardly ever get access to $g$ at all. With this in mind, we shrink the background region determined in the previous step to those pixels that map to the wall which appears most often in it.
\item\label{item:dltn} The homography between the image plane and the resulting background region, i.e., the CAVE wall that appears in most pixels and is seen directly, is computed by a standard direct linear transformation method. Given the projection matrix,  $g\in \SE(3)$ can be extracted from the associated homogeneous matrix~\cite{Zhang2000}.
\item We now revisit the full light map $l^*$ on the image plane. Every point $\bm{l}(\bm{p})$ that can be projected directly into the image plane, utilizing the initial estimate of $g$, must belong to the background. The characteristic function of the foreground then becomes
\[
\chi_{\mathrm{b}}(\bm{p}):=\begin{cases}1\quad \text{if} \;\|\pi(\mathbf{R}\bm{l}(\bm{p})+\bm{t})-\bm{p}\|<\theta, \\ 0\quad \text{otherwise},\end{cases}
\]
where $\pi$ denotes the canonical pinhole projection. Observe that the sensitivity of $\chi_b$ w.r.t. the threshold is quite low. In other words, the choice of $\theta$ requires no great care because the backprojection error grows discontinuously across the silhouette. The final result is depicted in Fig.~\ref{fig:sssegd}.
\end{enumerate}
With the final segmentation $\chi_b$, we improve the estimate of the extrinsic camera parameters $g$ by running a robust bundle adjustment on all points in the background~\cite{Zhang2000}. Let us remark that in previous experimental setups, the screen is rarely seen directly during measurement but obviously has to be for extrinsic calibration. To establish an oblique view on the screen, a calibration normal (usually a planar mirror) is necessary, which complicates Step~\ref{item:dltn} and the final bundle adjustment significantly~\cite{Knauer2004}. Also note that the procedure described in~\cite{Liu2011} is not applicable here as it requires the pattern generator to be movable. 
\begin{figure}[t]
\centering
\includegraphics[width=0.32\columnwidth]{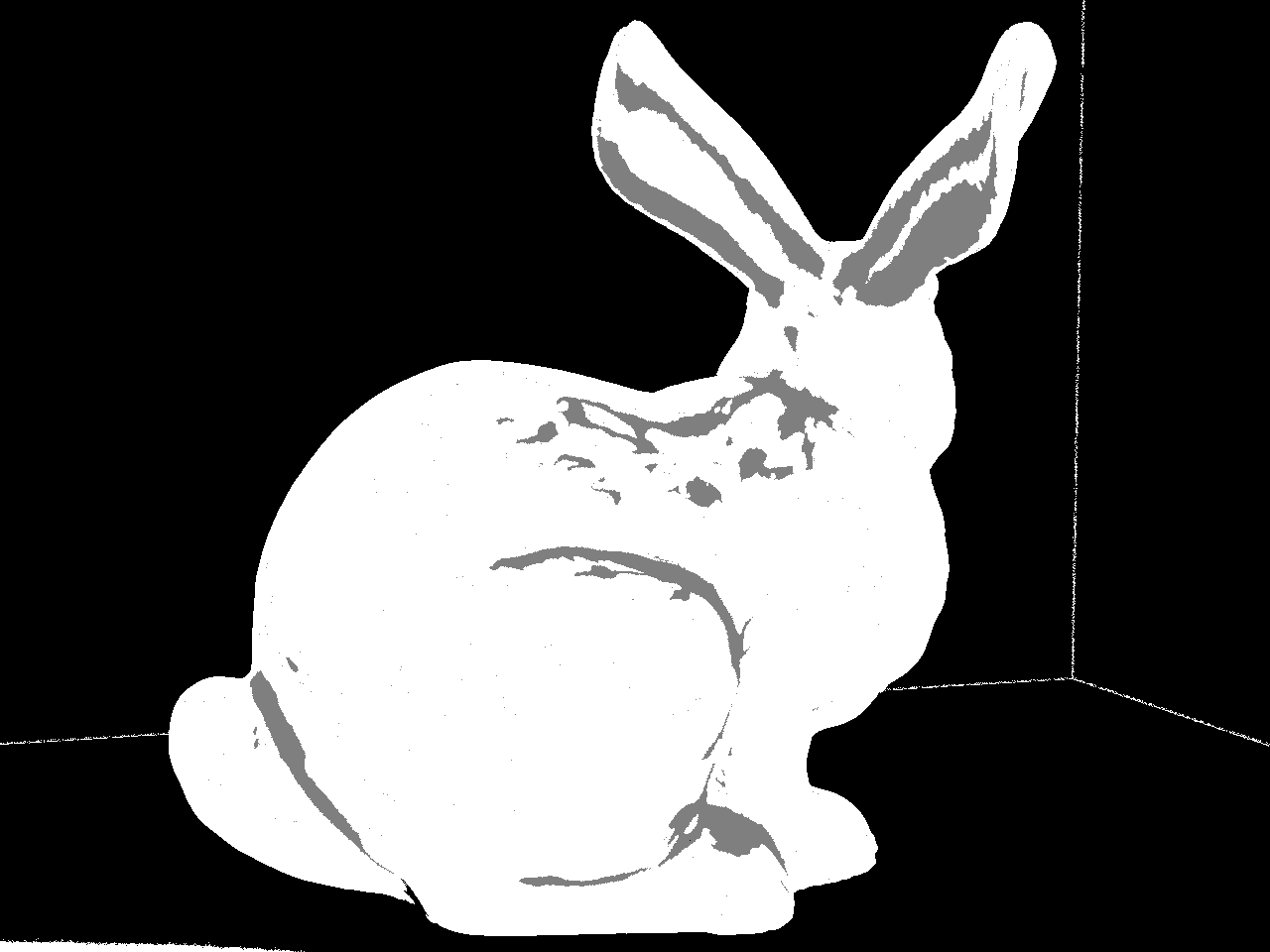}\hfill
\includegraphics[width=0.32\columnwidth]{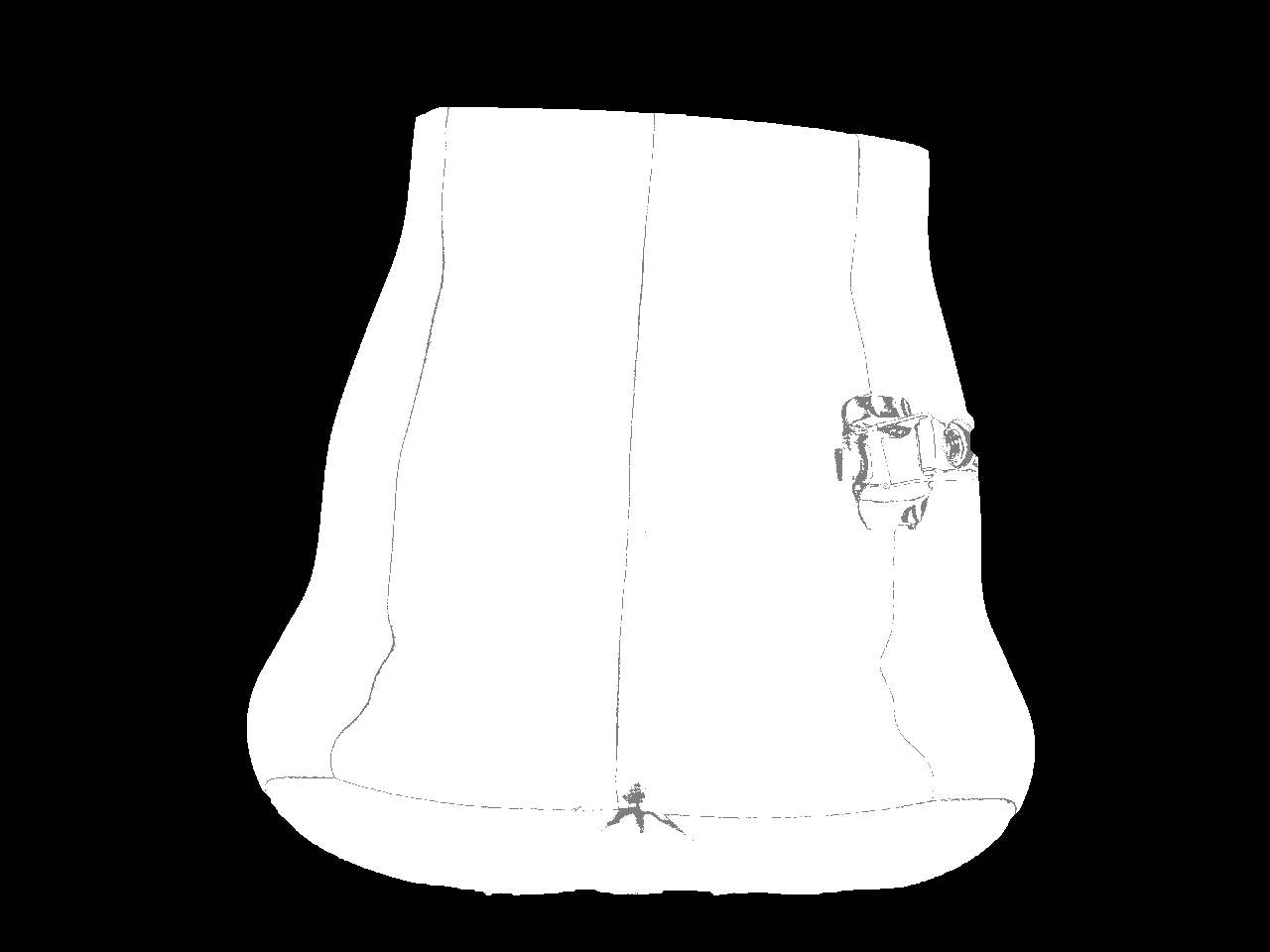}\hfill
\includegraphics[width=0.32\columnwidth]{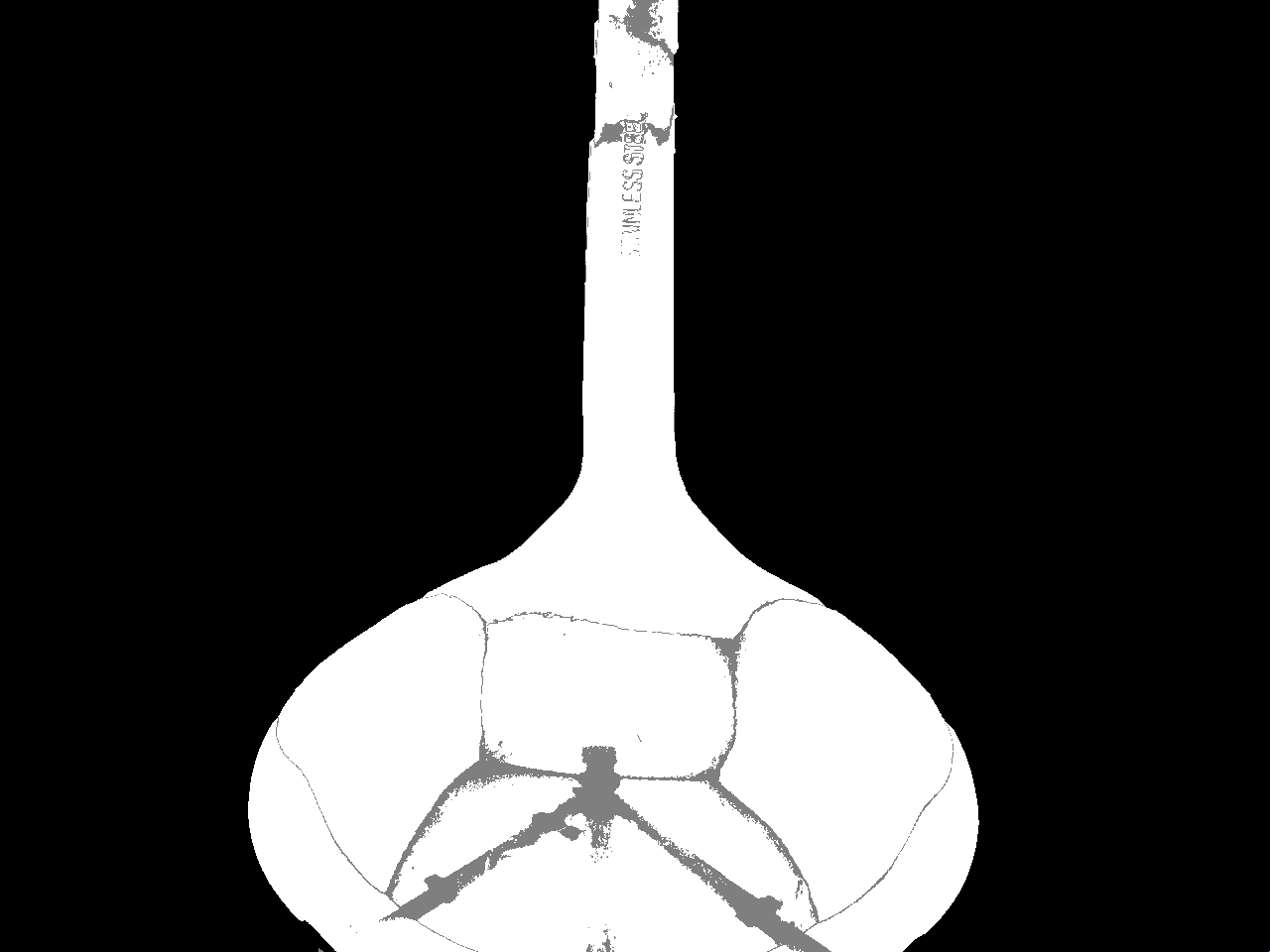}\vspace{0.05cm}\\
\includegraphics[width=0.32\columnwidth]{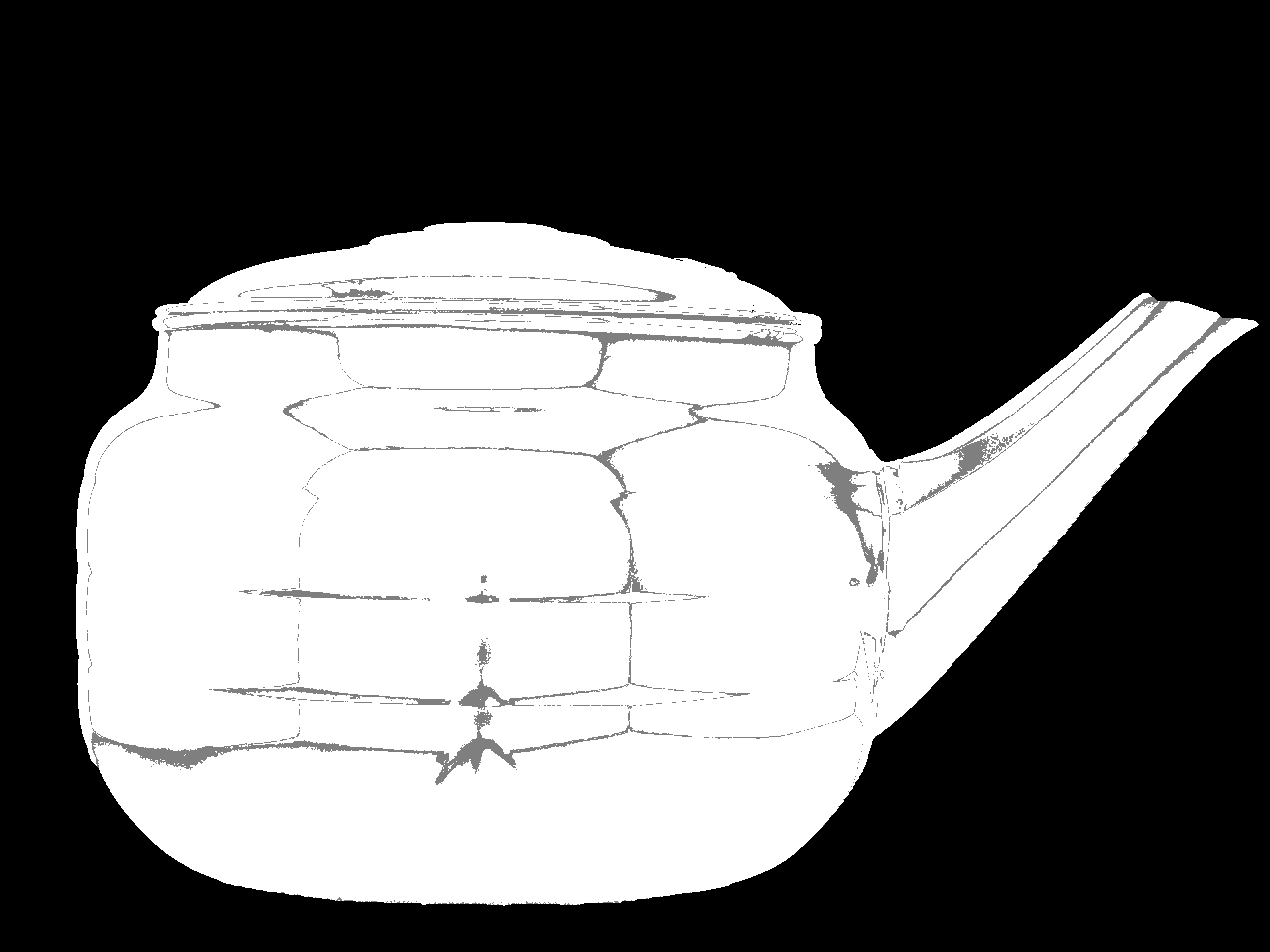}\hfill
\includegraphics[width=0.32\columnwidth]{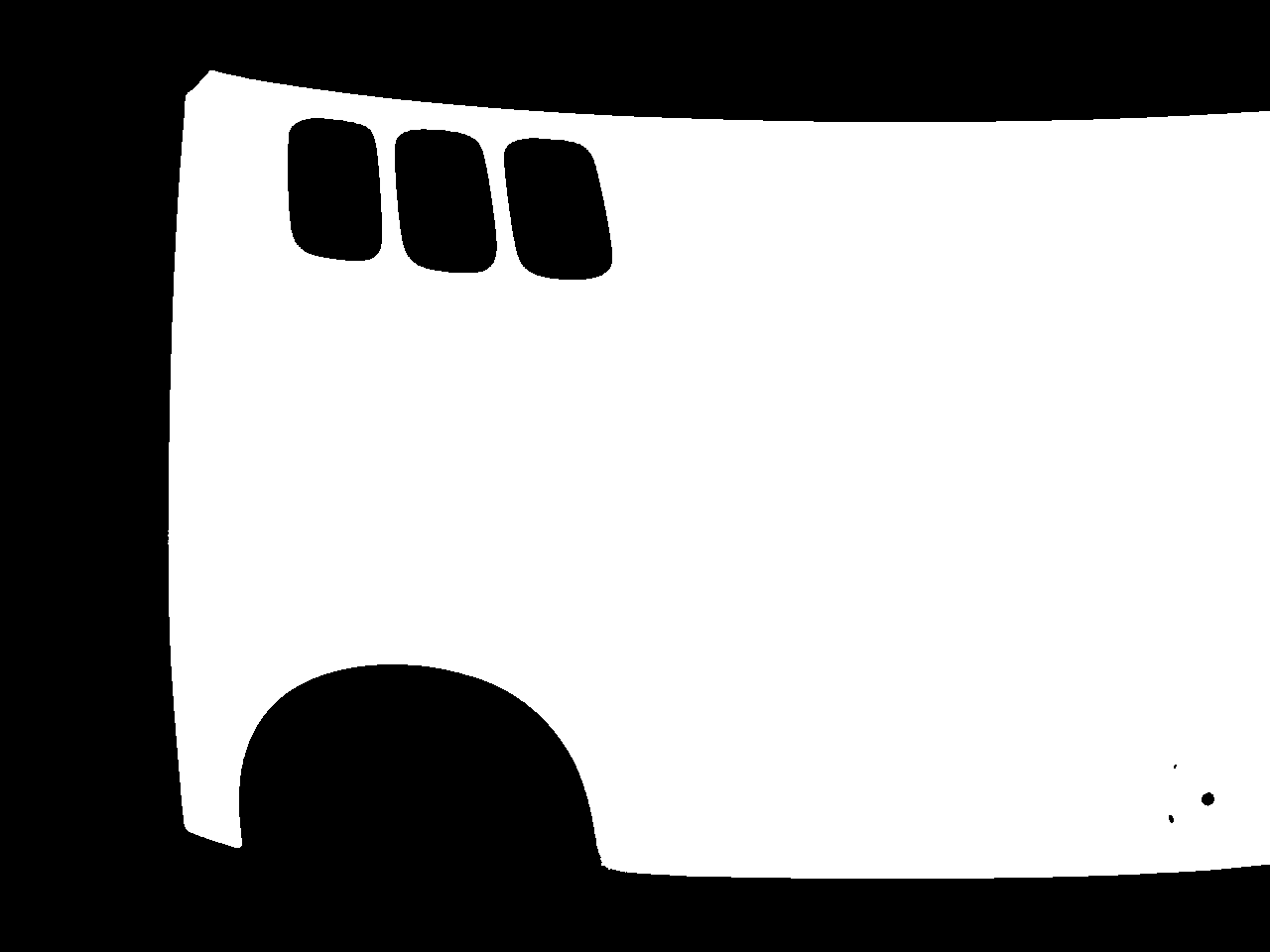}\hfill
\includegraphics[width=0.32\columnwidth]{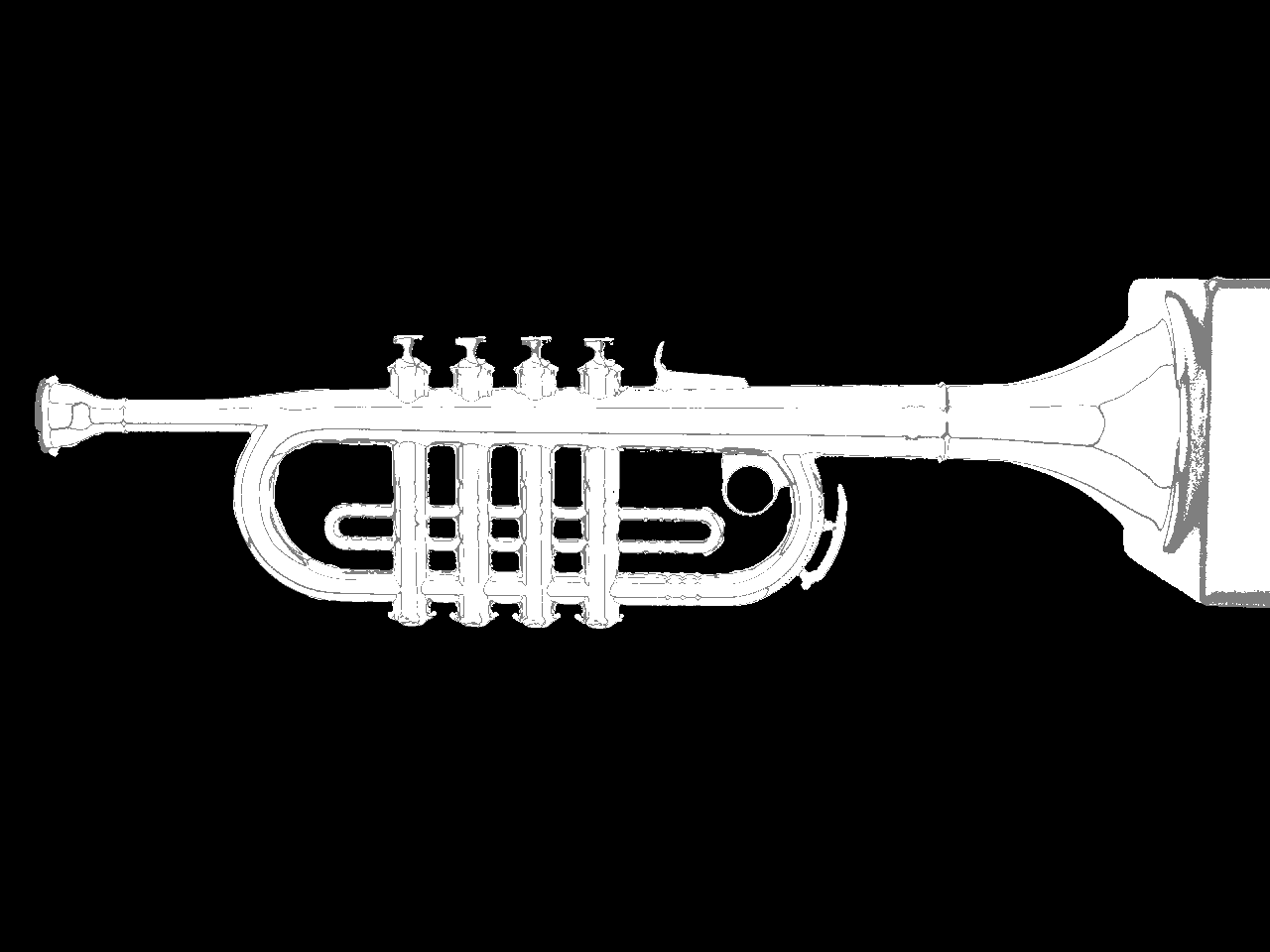}
\caption{Segmentation masks. Pixels with invalid measurements are shown in gray (e.g., on Lambertian parts of the object, the edges of the CAVE, or where the camera occludes the walls).}
\label{fig:masks}
\end{figure}
\section{Surface reconstruction}\label{sec:reconstruction}
What sets the deflectometric measurement principle apart from techniques such as stereopsis or laser scanning is that it yields \emph{slope} which must be converted into \emph{depth} by means of \emph{integration}. More precisely, the light map $l$ acquired from a vantage point $g$ is equivalent to a unit vector field of \emph{measured} normals $\nd(\bm{x})$ in space. To see this, fix a point $\bm{x}\in C$ inside the visible volume, transform it to camera coordinates via $g^{-1}\bm{x}$, project it to $S^2$ by nor\-ma\-li\-za\-tion, and look up the value of the light map $\bm{l}=l(\hat{\bm{x}})$ (Fig.~\ref{fig:geometry}). Suppose that $\bm{x}\in S$, then the surface normal $\nn(\bm{x})$ must obey the law of reflection, i.e., the correspondence between $\bm{x}$ and $\bm{l}$ yields a hypothesis $\nd$ about the normal the true surface should have if it indeed contained the point $\bm{x}$. We say that a regular surface $S$ solves the reconstruction problem if it interpolates the measured normal field in the sense that $\nn(S)=\nd|_{S}$ except on zero-measure subsets of $S$. Such an $S$ rarely exists due to stochastic disturbances of the data. A maximum-likelihood estimate of the surface is obtain as the solution of the following variational problem:
\begin{equation}\label{eq:energy}
	S^*=\arg\min_S\int\limits_{C}\frac{1}{2}\|\hat{\bm{n}}(\bm{x})-\nd(\bm{x})\|^2\,\mathrm{d}\bm{x}.
\end{equation}
The well-known gauge ambiguity manifests itself in the dependence of $\nd$ on $\bm{x}$: an additional constraint $\bm{x}_0\in S$ needs to be imposed to render problem~\eqref{eq:energy} well-posed~\cite{Balzer2010}. Since in this paper, we are mainly concerned with the deflectometric data acquisition process and less with reconstruction, we assume the point $\bm{x}_0$ to be known for now. The next section will outline a suitable initialization heuristic. For more sophisticated regularization approaches, we refer the interested reader to the related literature, see Sect.~\ref{subsec:relatedwork}. Also, at this point, we make no attempt to fuse data acquired from a series of vantage points into a common normal field. Our data could be processed with any existing multiview normal field integration scheme, cf.~\cite{Bonfort2006,Weinmann2013}.
\section{Experimental evaluation}
\begin{figure}[t]
\centering
\includegraphics[width=0.8\columnwidth]{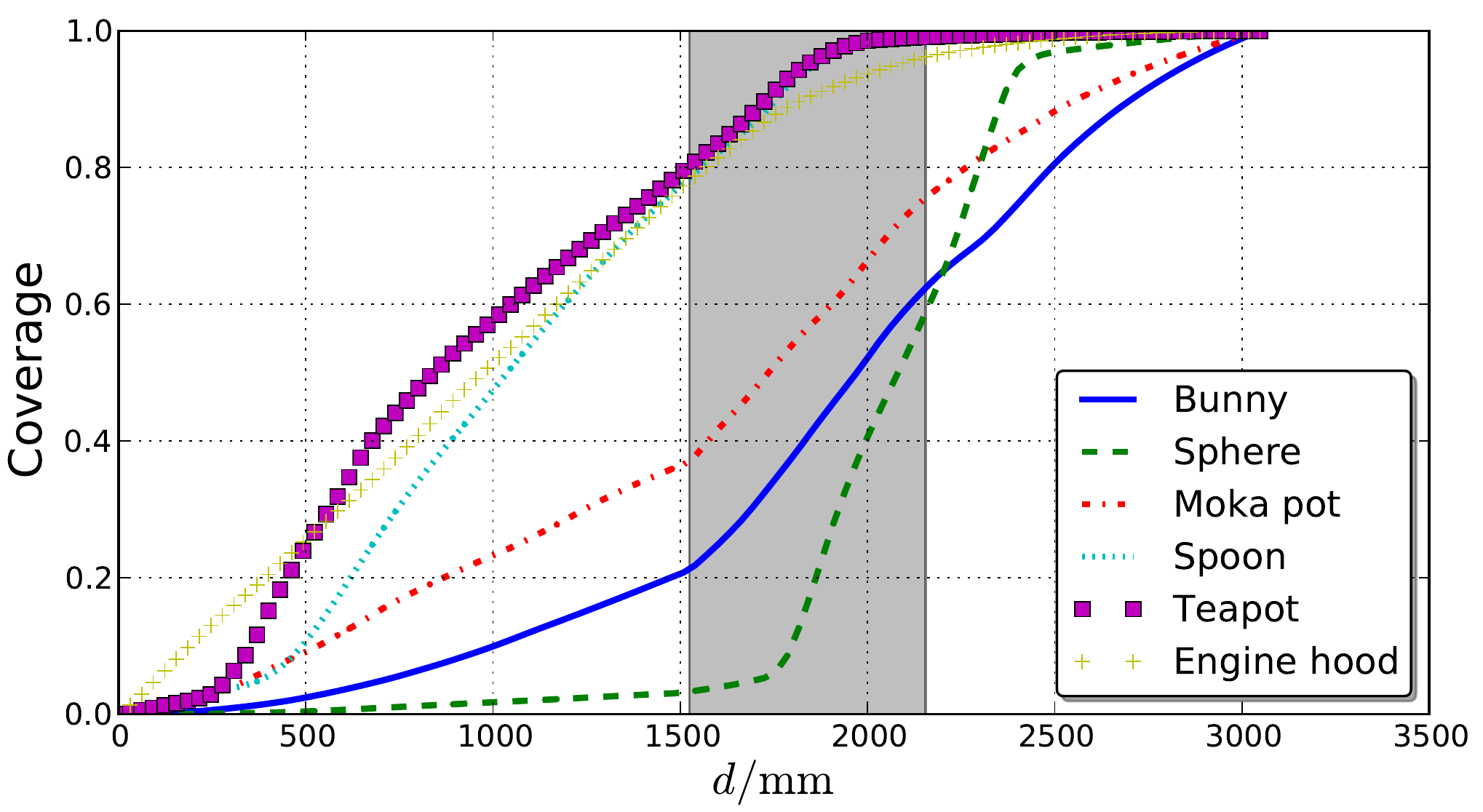}
\caption{Ratio of the area covered by projections of surface points on the image plane (in percentage of pixels) over the area on the screen (parametrized by the distance $d$ to a fixed point) which is seen through these points. Focal length: $16.3$~$\mathrm{mm}$.}\label{fig:coverage}
\end{figure}
\subsection{Implementation}\label{subsec:implementation}
\begin{figure}[b]
\centering
\subfigure[$k=0$]{\includegraphics[width=0.23\columnwidth]{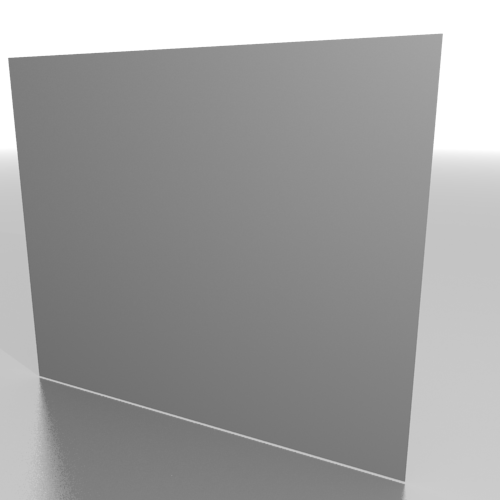}}\hfill
\subfigure[$k=1$]{\includegraphics[width=0.23\columnwidth]{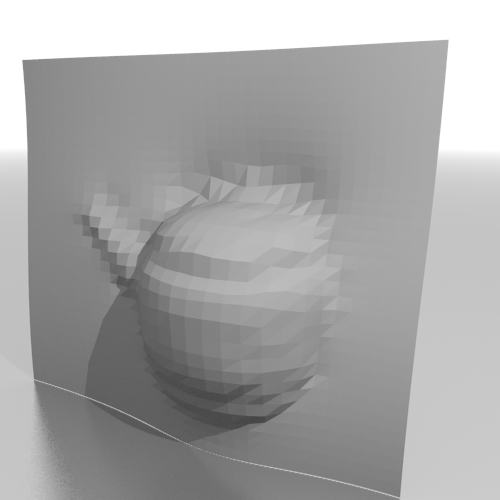}}\hfill
\subfigure[$k=5$]{\includegraphics[width=0.23\columnwidth]{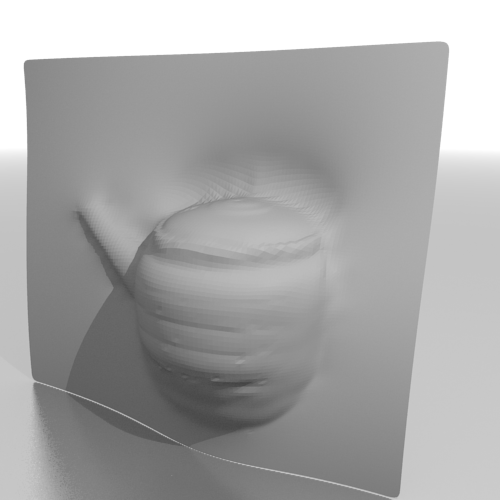}}\hfill
\subfigure[$k=11$]{\includegraphics[width=0.23\columnwidth]{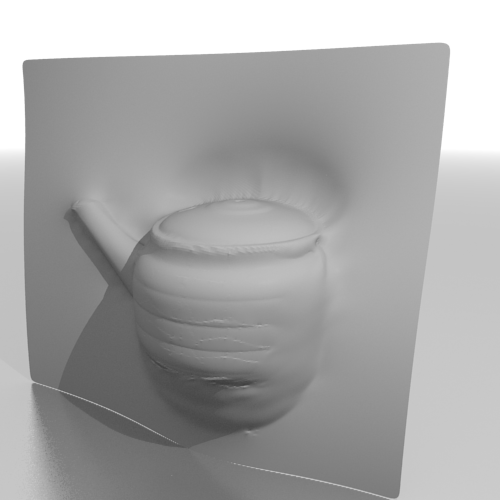}}
\caption{Four iterations of the normal integration method with intermediate uniform refinements of the triangular mesh.}
\label{fig:intiter}
\end{figure}

We implemented most of the pipeline in C++, except for the steps related to optical coding. All experiments were run on a single $3.4$~$\mathrm{GHz}$ core of a commodity computer. We solve the shape optimization problem~\eqref{eq:energy} with the algorithm proposed in~\cite{Balzer2012}. Since integration is of broader interest, we will make the source code available. As an initial guess, we use a fronto-parallel plane which is also tangent to the visual hull. Loop subdivision is applied to the evolving mesh between two steps to increase the level of detail and prevent the iteration from falling into local minima of the cost functional\footnote{although this is less critical given that the method is of second order}, see Fig.~\ref{fig:intiter}.

\subsection{Results}\label{sec:results}
A selection of segmentation masks obtained with the method developed in Section~\ref{sec:calibration} is shown in Fig.~\ref{fig:masks}. The computation of a single mask constitutes the main bottleneck requiring in average $15$ $\mathrm{s}$ on a single CPU core. Thereby, the majority of time is consumed by the vast number of singular value decompositions governing the rank estimation in~\eqref{eq:scoringfunction}. The following experiment investigates the impact of screen enlargement: From the pose previously obtained by calibration, the barycenter of the CAVE wall can be determined which is located \emph{behind} the camera. Fig.~\ref{fig:coverage} shows the number of points in the image of different light maps $l$ closer than a distance $d$ to that barycenter. We limit $d$ to $2h$, as any point further away is likely a part of the background. The shaded area corresponds to the walls next to, above, or below the object, and it can be observed that these may contribute significantly to the number of valid measurements among the pixels of a single deflectometric image. In contrast, the monitor of a classical setup would only occupy a small portion of the back wall. 

\begin{figure}[t]
\centering
\def \myfigwidth{0.45\columnwidth}
\subfigure[Bunny]{\label{fig:bunny}%
\begin{tabular}[b]{c}
\includegraphics[width=\myfigwidth]{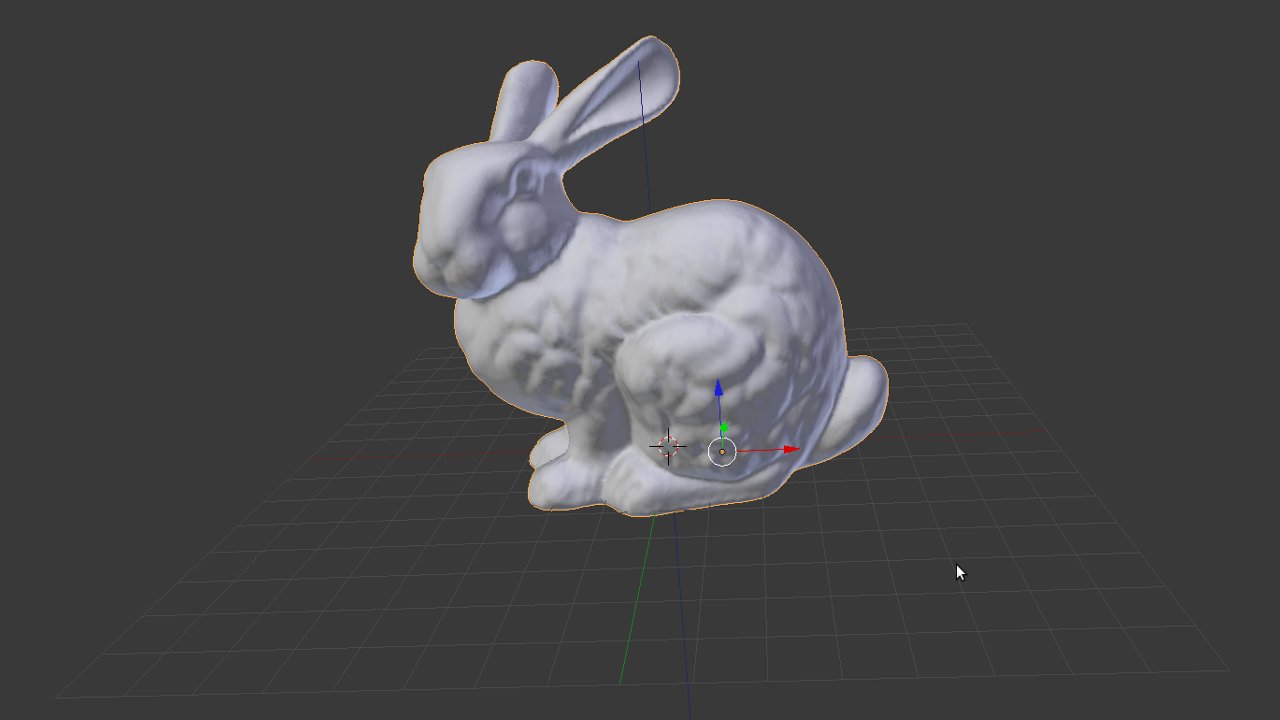}\\
{\def\svgwidth{\myfigwidth}\tiny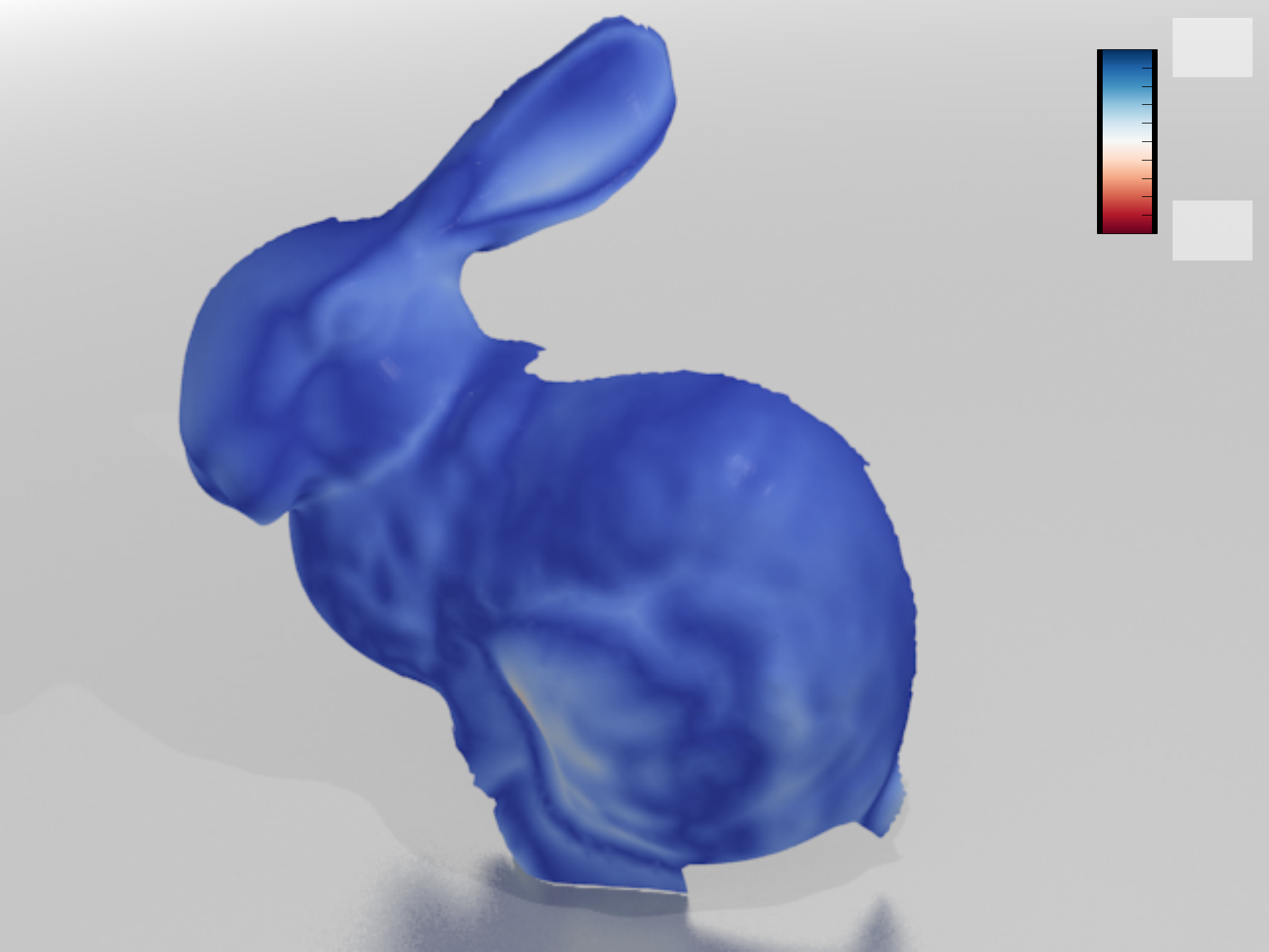}
\end{tabular}}%
\subfigure[Moka pot]{\label{fig:moka}%
\begin{tabular}[b]{c}
\includegraphics[width=\myfigwidth]{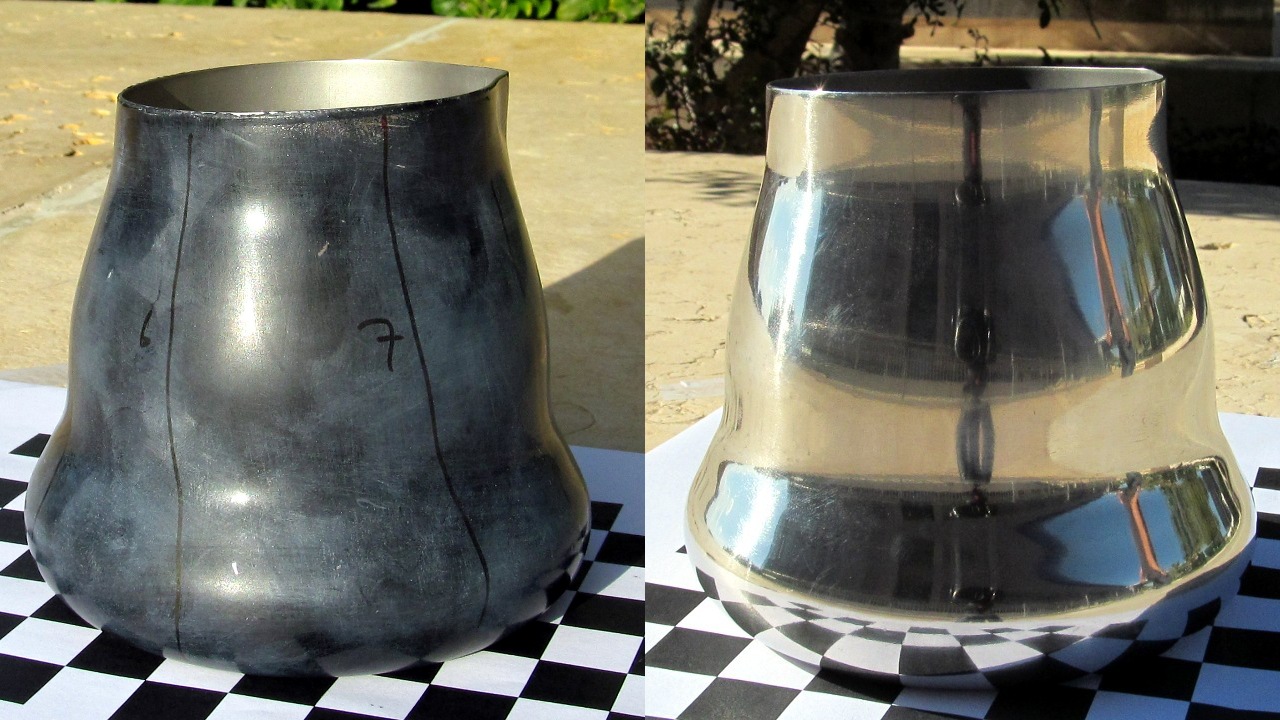}\\
{\def\svgwidth{\myfigwidth}\tiny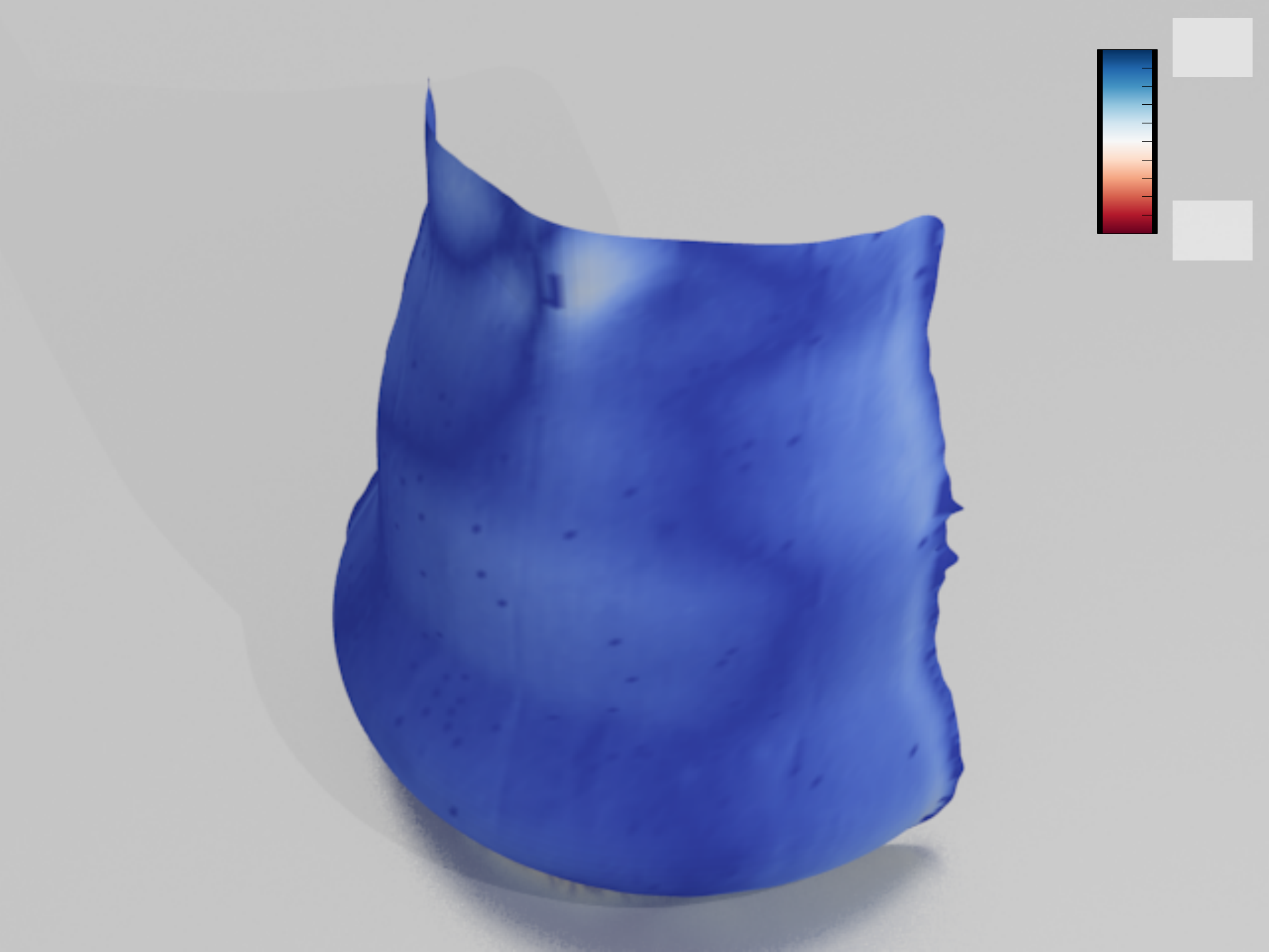}
\end{tabular}}\\
\subfigure[Teapot]{\label{fig:teapot}%
\begin{tabular}[b]{c}
\includegraphics[width=\myfigwidth]{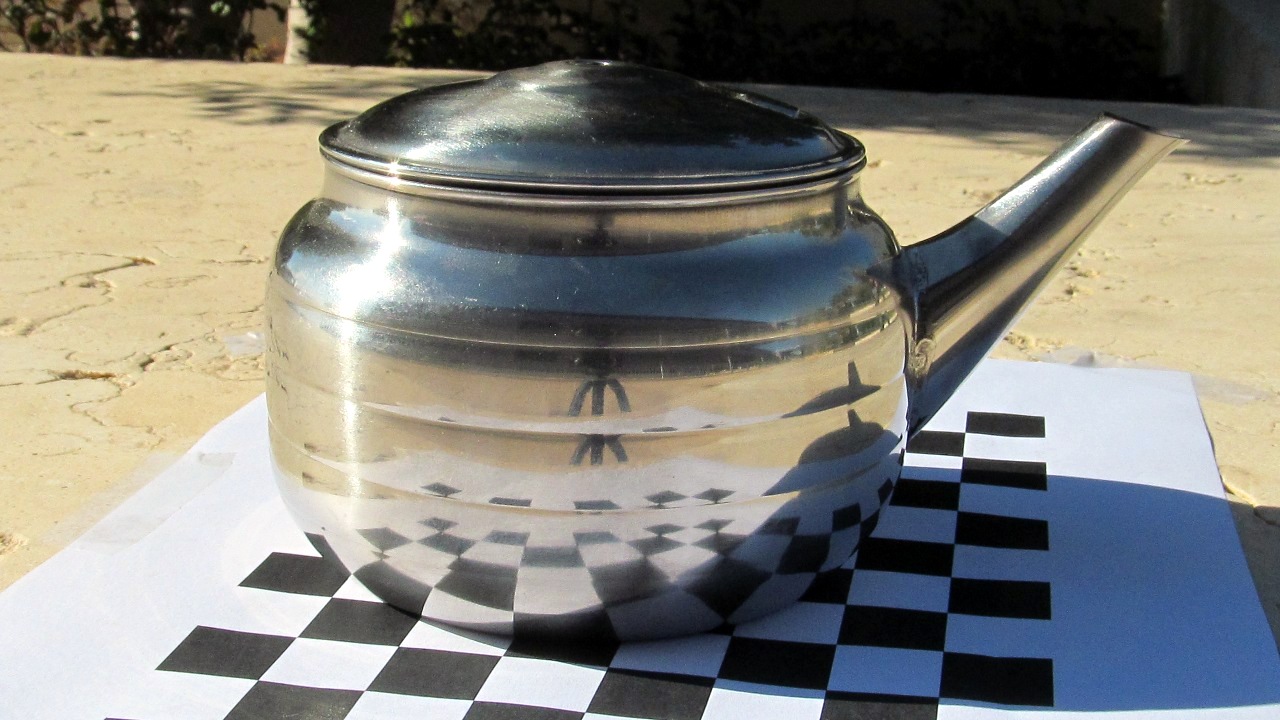}\\
{\def\svgwidth{\myfigwidth}\tiny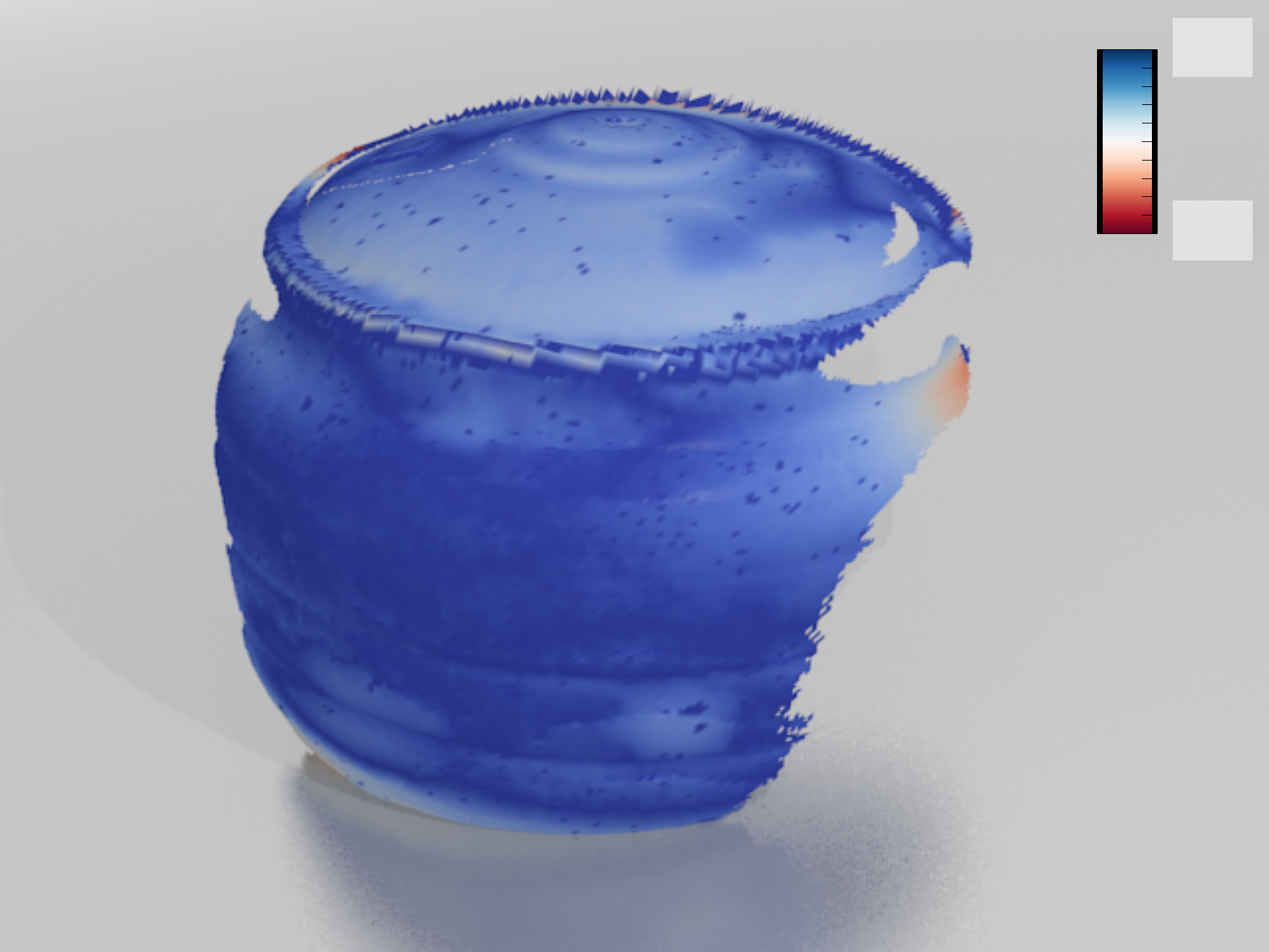}
\end{tabular}}%
\subfigure[Engine hood]{\label{fig:hood}%
\begin{tabular}[b]{c}
\includegraphics[width=\myfigwidth]{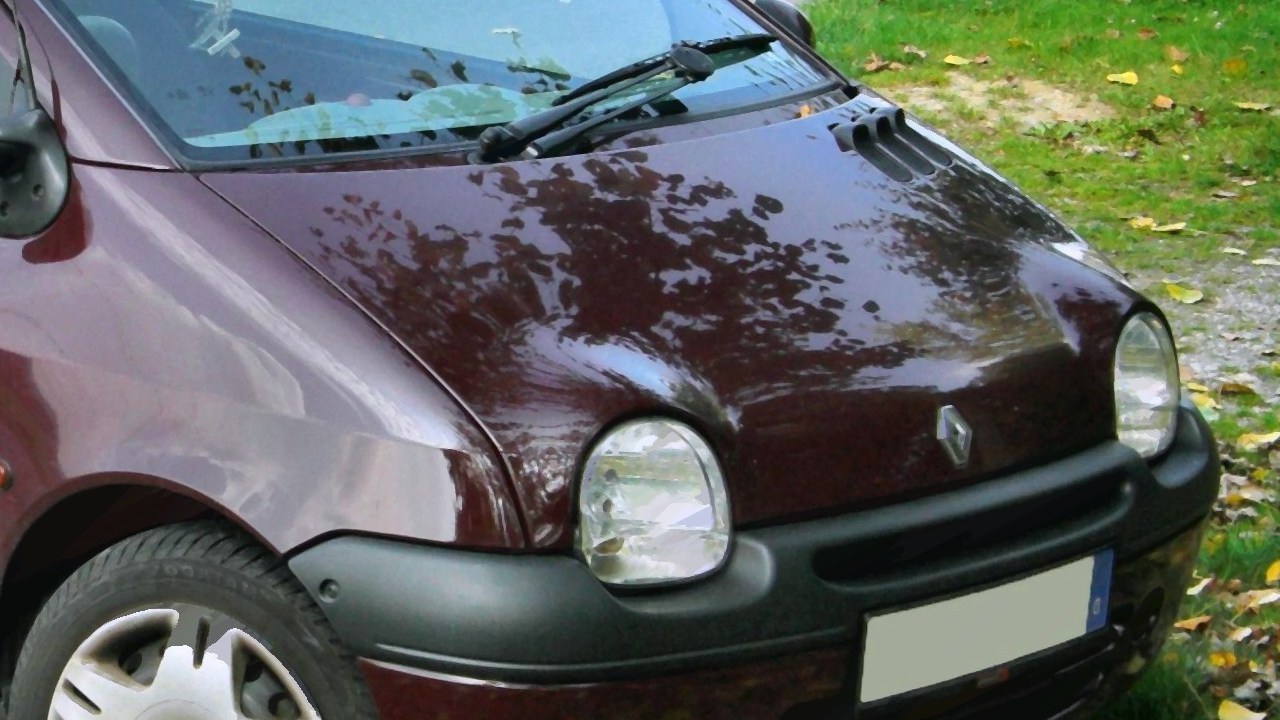}\\
{\def\svgwidth{\myfigwidth}\tiny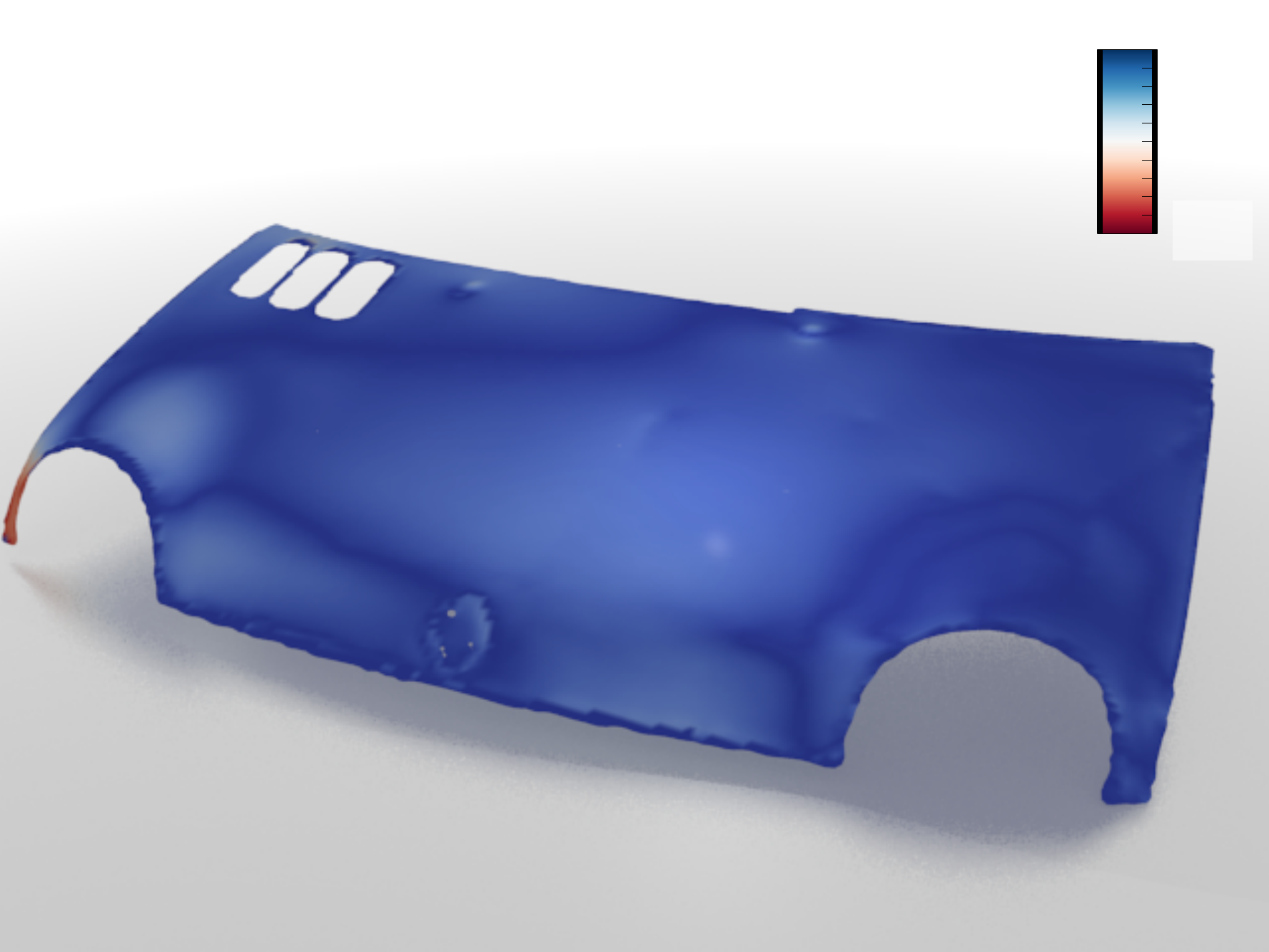}
\end{tabular}}%
\caption{A picture of the specimen and our reconstruction colored by local deviation from ground truth in $\mathrm{mm}$.}
\label{fig:finalresults}
\end{figure}

Our experimental setup is unique in the realm of deflectometry which makes the comparison with other methods challenging. For this reason, we painted some of the specimens black and coated them with white powder, which is very undesirable in practical applications but unavoidable when using a laser scanner as an alternative source of data (Figs.~\ref{fig:moka} top left). Needless to say that range scans are afflicted with uncertainty of their own. Perfect ground truth is only known after simulating data by ray tracing (Fig.~\ref{fig:bunny}) or in the form of a CAD model supplied by the manufacturer (Fig.~\ref{fig:hood}). After aligning reconstructed and ground-truth model by the iterative closest point method, the local deviation between the two becomes apparent. Its magnitude is foreshadowed by the vertex colors of the meshes depicted in Fig.~\ref{fig:finalresults}. In all examples, the error peaks close to the occluding boundary where the actual surface normal is almost perpendicular to the image plane. The mean error is determined by the resolutions of camera and the projectors but also their geometric distance from the surface. The latter is maximized towards better coverage here, which puts the mean error in the range one would expect from a back-of-the-envelope estimate of the backprojected pixel size. Let us remark that our system could be operated in a hierarchical way, where a holistic but coarse reconstruction is examined for abnormalities first, followed by a high-precision close-up scan of conspicuous regions. Aggregate performance statistics are summarized in Tab.~\ref{tab:performance}: Note that \emph{all reconstructions stem from a single capture}. In comparison, Weinmann et al.~\cite{Weinmann2013} reported sample sizes ranging from $500$ to $750$. We were also able to reduce the number of measurements for the same type of engine hood investigated in \cite{Balzer2011} from $265$ to $1$. 

\begin{figure}[t]
\centering
\def \myfigwidth{0.3\columnwidth}
\subfigure[Moka pot]{\label{fig:moka_render}%
\begin{tabular}[b]{c}
\includegraphics[width=\myfigwidth]{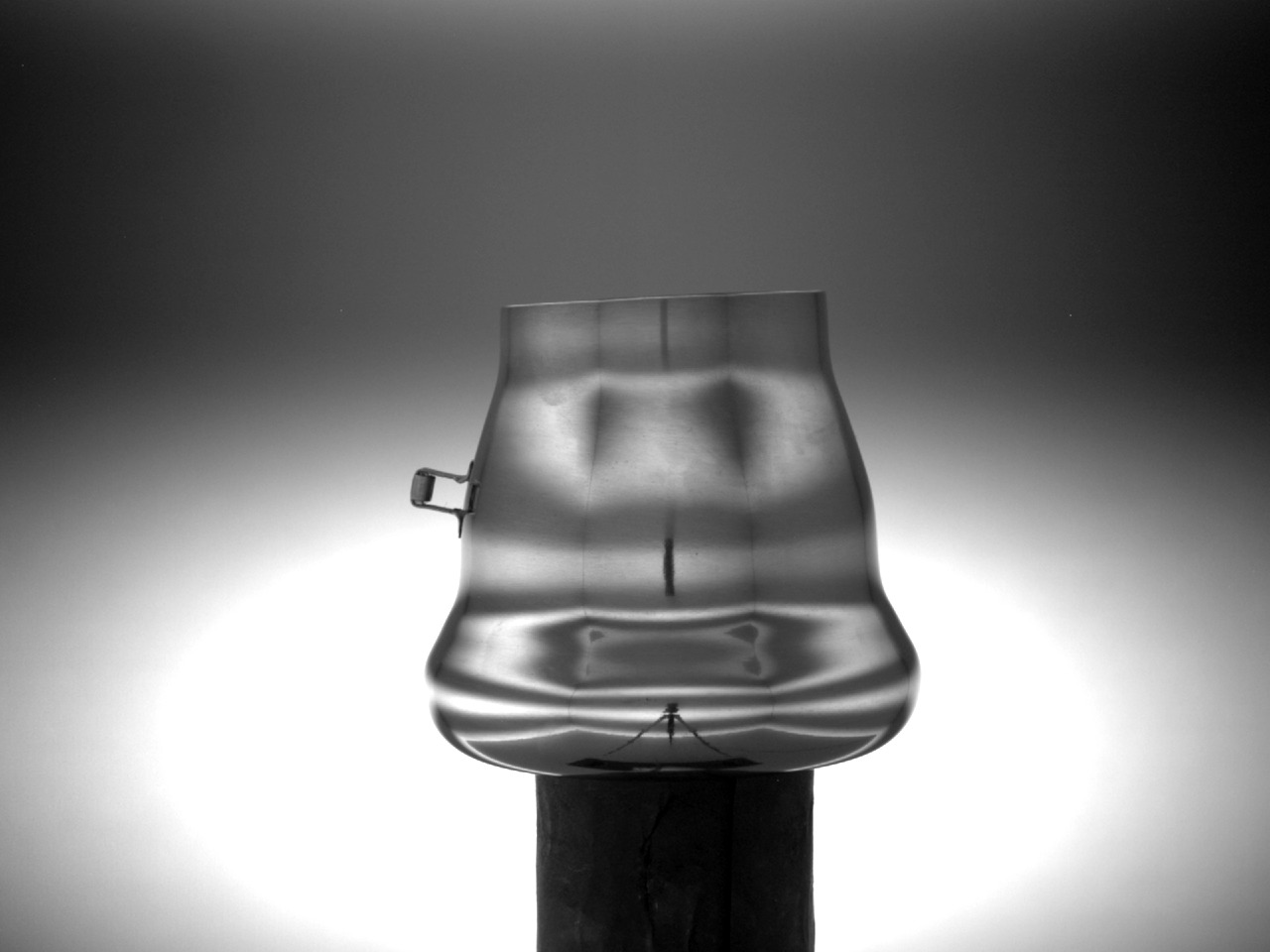}\\
\includegraphics[width=\myfigwidth]{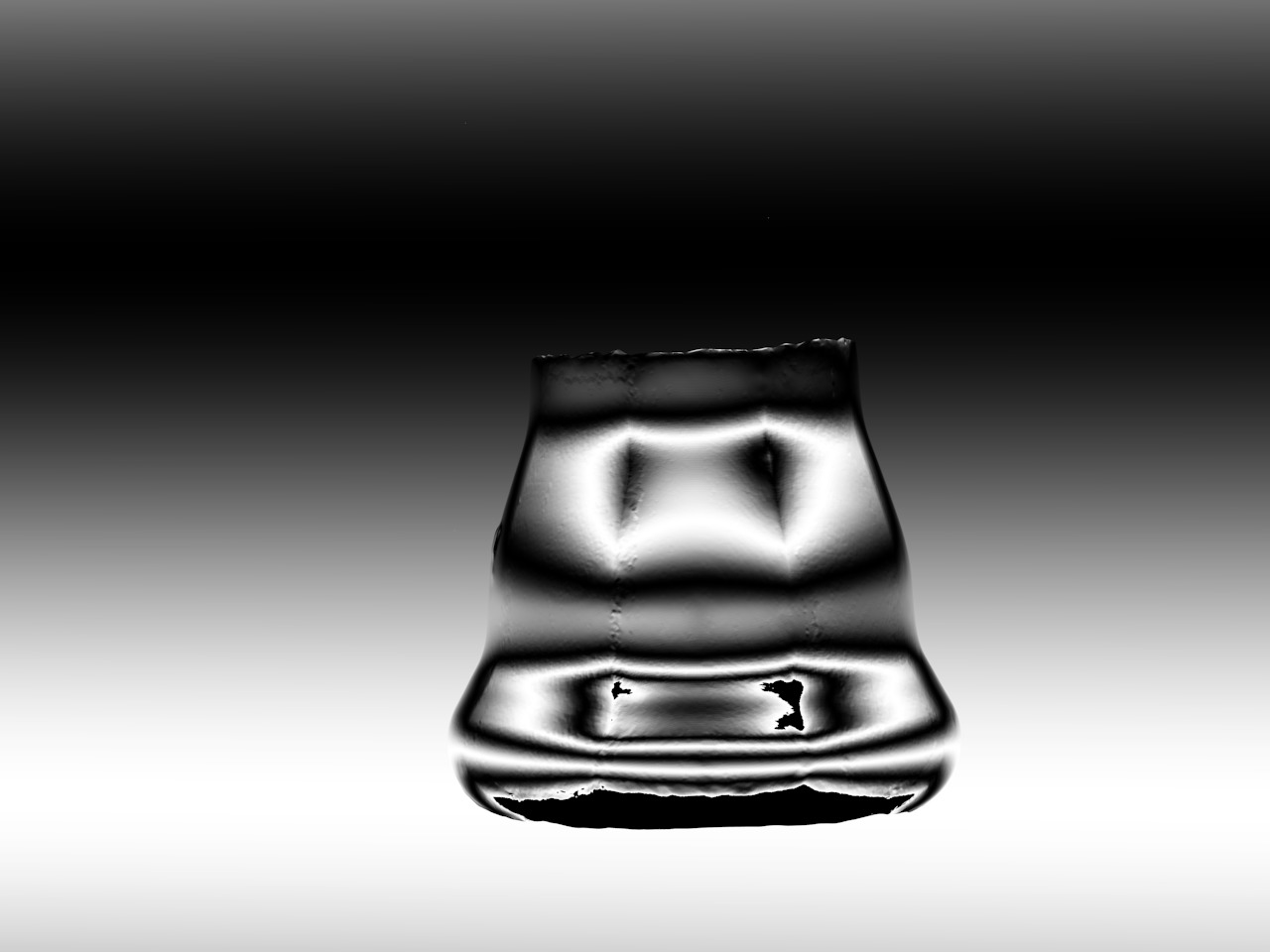}
\end{tabular}}%
\subfigure[Teapot]{\label{fig:teapot_render}%
\begin{tabular}[b]{c}
\includegraphics[width=\myfigwidth]{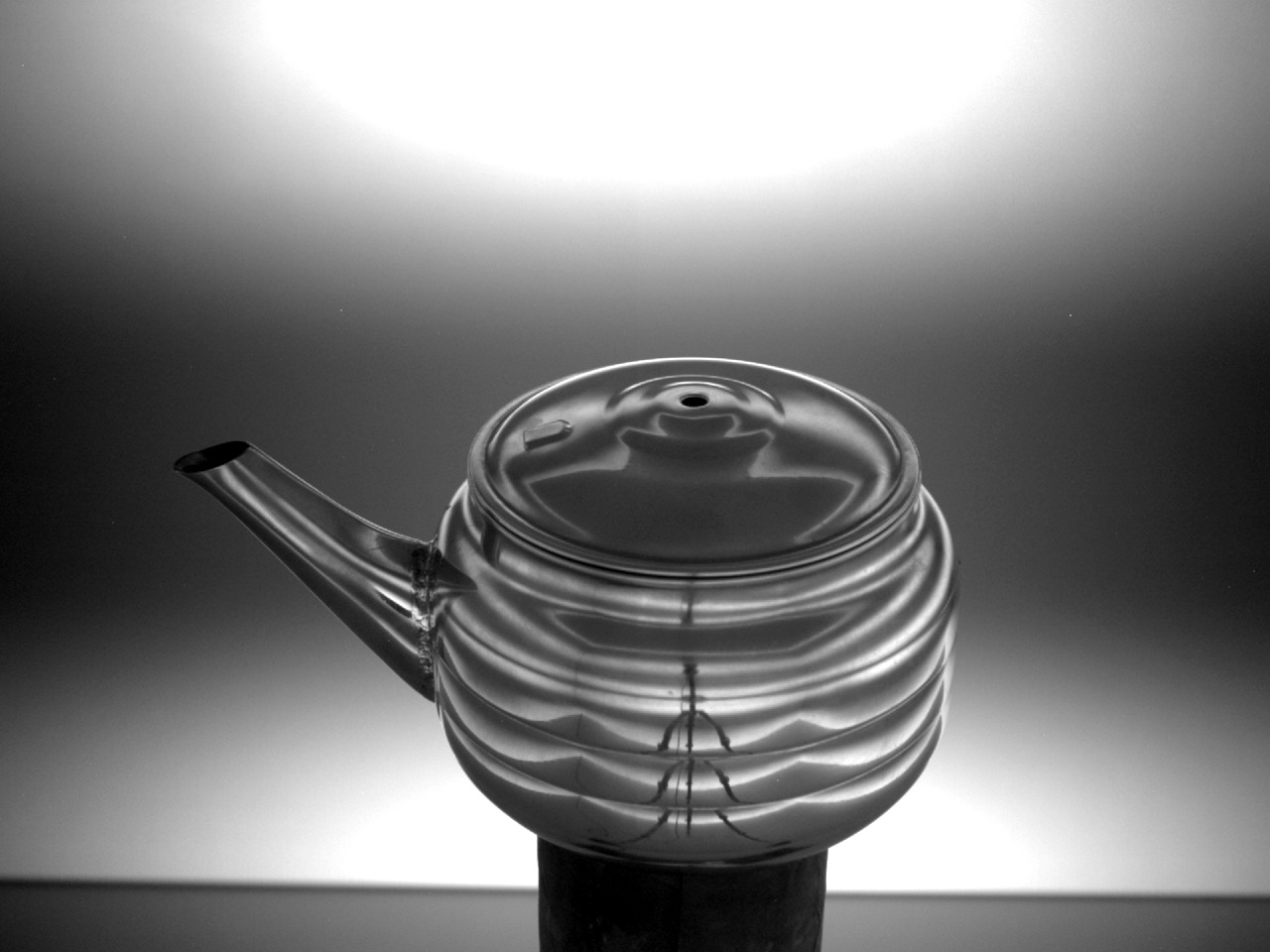}\\
\includegraphics[width=\myfigwidth]{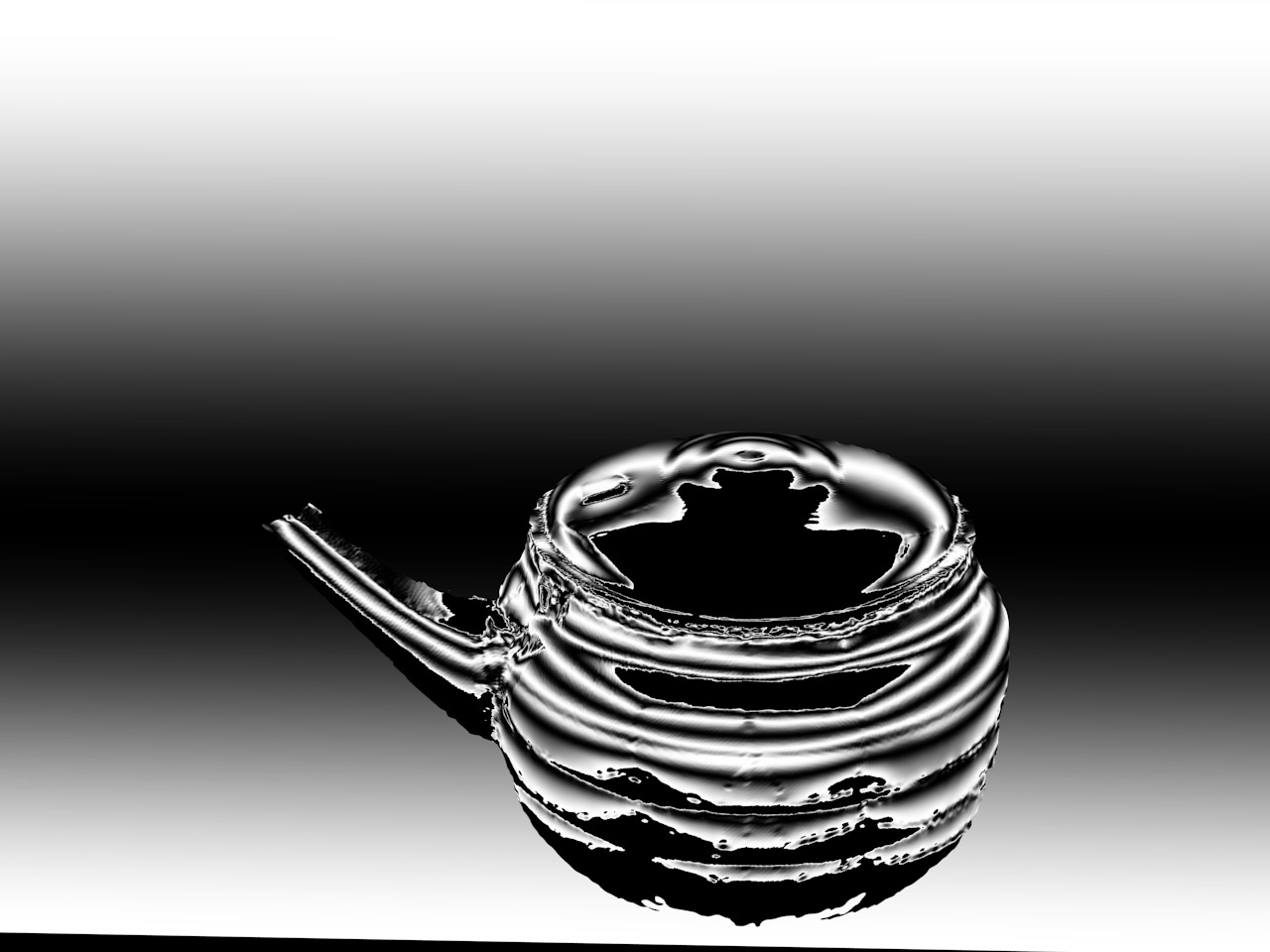}
\end{tabular}}%
\subfigure[Engine hood]{\label{fig:hood_render}%
\begin{tabular}[b]{c}
\includegraphics[width=\myfigwidth]{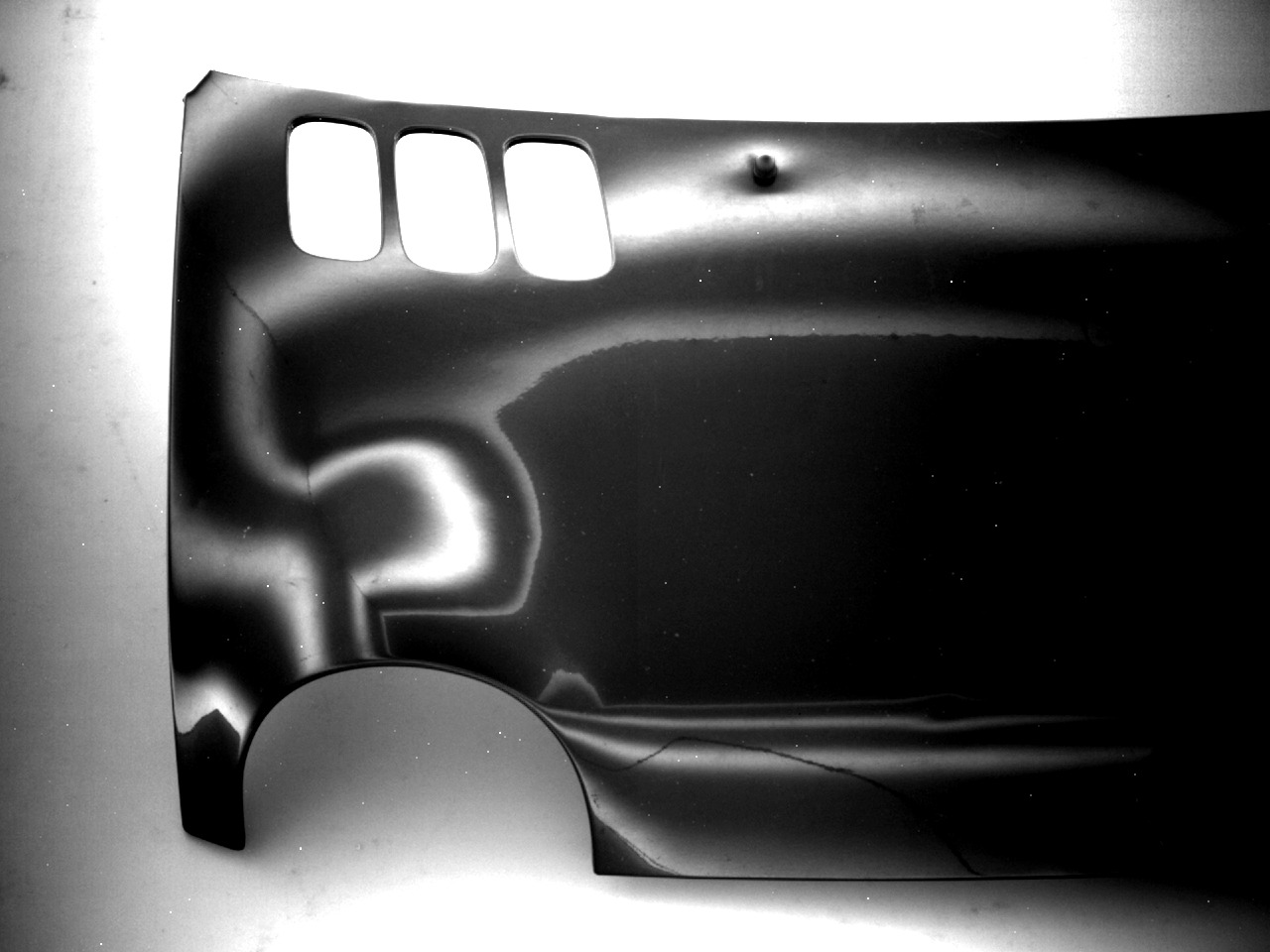}\\
\includegraphics[width=\myfigwidth]{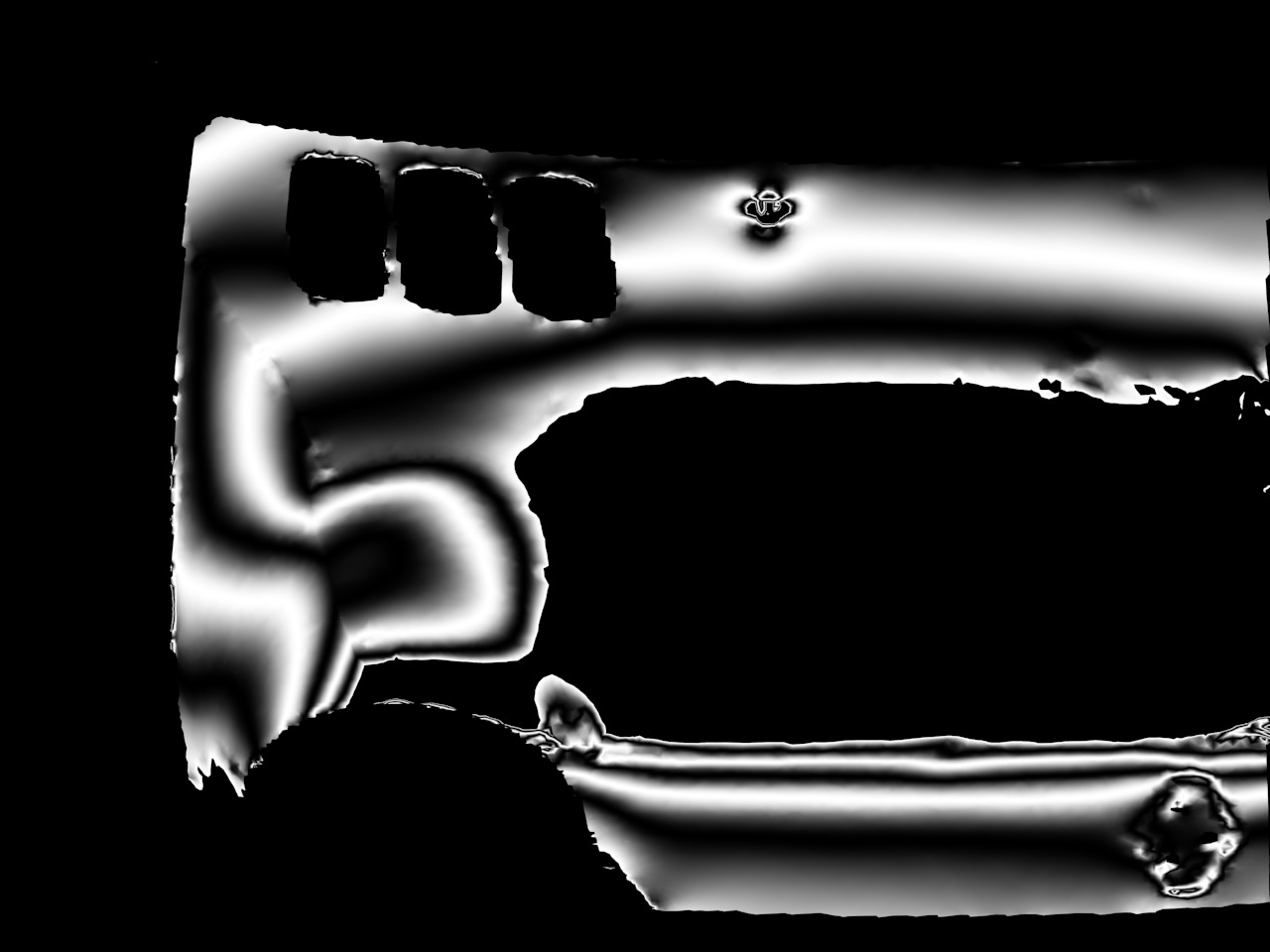}
\end{tabular}}%
\caption{Closing the loop: Real images acquired in the CAVE (top) compared to simulations (bottom) based on vantage points and object models obtained by our method.}
\label{fig:resulteval}
\end{figure}

Finally, we used the recovered meshes and vantage points to render images of the scene they constitute together with a 3-d model of the CAVE. As shown in Fig.~\ref{fig:resulteval}, the image series leading to the underlying reconstructions could be reproduced with satisfactory accuracy. 

\begin{table}[tb]
\centering
 \setlength{\tabcolsep}{3pt}
\def \mycolwidth{1.5cm}
\footnotesize
\begin{tabular}{lccccc}\toprule
\bf{Object} & $\text{\bf No. faces}\atop ~$ & $\text{\bf No. views}\atop ~$ & $\text{\bf Area}\atop [m^2]$ & $\text{\bf Mean error}\atop[mm]$ &  $\text{\bf Max. error}\atop[mm]$  \\\otoprule %
\emph{Bunny} & 26681 & 1 & 0.57 & 0.002 & 0.594  \\\midrule
\emph{Moka pot} & 25630 & 1 & 0.015 & 0.27 & 3.69 \\\midrule 
\emph{Teapot} & 61922 & 1 & 0.011 & 0.39 & 2.8 \\\midrule
\emph{Hood} & 44588 & 1 & 0.76 & 1.12 & 13.6\\\otoprule
\end{tabular}
\caption{Performance statistics.}
\label{tab:performance}
\end{table}

\section{Conclusion}
In this paper, we have argued that deflectometric imaging and reconstruction benefit strongly from illuminants that encompass all of the scene except for the mirror object under inspection. These can be realized e.g. by means of a CAVE. While the cost of building and maintaining such a sensor system may seem limiting, the proposed algorithms transfer -- mutatis mutandis -- to testbeds far less expensive yet equally effective. The automotive industry, e.g., employs \emph{light tunnels} albeit not for reconstruction. A low-cost alternative in a regular room with opaque white walls and cheap standard projectors is shown in Fig.~\ref{fig:microcave}. In conclusion, let us remark that the CAVE, beyond its role as deflectometry illuminant, may be utilized for reconstruction in general. To our best knowledge, it has been exclusive to visualization and virtual reality so far, but in our opinion, it opens up possibilities in several other topics of computer vision and optical metrology.

\bibliographystyle{alpha}

\end{document}